\documentclass{article}
\usepackage{graphicx}
\usepackage{mathptmx}      
\usepackage{amsmath}
\usepackage{svg}
\usepackage{tcolorbox}
\usepackage{amssymb}
\usepackage{graphicx}
\usepackage{prodint}
\usepackage{rotating}
\usepackage{doi}
\usepackage{bbm}
\usepackage{paralist}

\usepackage{booktabs}
\usepackage{longtable}
\usepackage{array}
\usepackage{multirow}
\usepackage{wrapfig}
\usepackage{float}
\usepackage{colortbl}
\usepackage{pdflscape}
\usepackage{natbib}
\usepackage{placeins}

\usepackage[compatibility=false]{caption}
\usepackage{subcaption}

\usepackage{authblk}
\usepackage[utf8]{inputenc}
\usepackage{bm}
\usepackage{tikz}
\usepackage{hyperref}
\usepackage{enumitem}

\usepackage[margin=1in]{geometry}

\title{Reduction Techniques for Survival Analysis}
\author[1,3]{Johannes Piller\textsuperscript{*}} 
\author[4]{L\'ea Orsini\textsuperscript{*}} 
\author[1,5]{Simon Wiegrebe}
\author[8,9]{John Zobolas}
\author[3,6,7]{Lukas Burk}
\author[6]{Sophie Hanna Langbein}
\author[3]{Philip Studener}
\author[3]{Markus Goeswein}

\author[2,3]{Andreas Bender}
\affil[1]{Statistical Consulting Unit StaBLab, Department of Statistics, LMU Munich, Munich, Germany.}
\affil[2]{Department of Statistics, LMU Munich, Munich, Germany.}
\affil[3]{Munich Center for Machine Learning, LMU Munich, Munich, Germany.}
\affil[4]{Oncostat U1018, Inserm, labeled Ligue Contre le Cancer, University Paris-Saclay, 114 rue Edouard Vaillant, 94800, Villejuif, France}
\affil[5]{Department of Genetic Epidemiology, University of Regensburg, Regensburg, Germany.}
\affil[6]{Leibniz Institute for Prevention Research and Epidemiology – BIPS, 28359 Bremen, Achterstraße 30, Germany}
\affil[7]{Faculty of Mathematics and Computer Science, University of Bremen, Bibliothekstr. 1, 28359 Bremen}
\affil[8]{Department of Cancer Genetics, Institute for Cancer Research, Oslo University Hospital (OUS)}
\affil[9]{Department of Biostatistics, Oslo Centre for Biostatistics and Epidemiology (OCBE),  University of Oslo (UiO), 0379 Oslo, Ullernchausseen 64-66, Norway}
\date{}                     
\title{Reduction Techniques for Survival Analysis}

\begin{document}

\maketitle

\begingroup
\renewcommand\thefootnote{*}
\footnotetext{These authors contributed equally to this work.}
\endgroup

\begin{abstract}
\noindent
In this work, we discuss what we refer to as reduction techniques for survival analysis, that is, techniques that "reduce" a survival task to a more common regression or classification task, without ignoring the specifics of survival data. 
Such techniques particularly facilitate machine learning-based survival analysis, as they allow for applying standard tools from machine and deep learning to many survival tasks without requiring custom learners. 
We provide an overview of different reduction techniques and discuss their respective strengths and weaknesses. 
We also provide a principled implementation of some of these reductions, such that they are directly available within standard machine learning workflows.
We illustrate each reduction using dedicated examples and perform a benchmark analysis that compares their predictive performance to established machine learning methods for survival analysis.

\vspace{1em}
\end{abstract}

\section{Introduction}

Survival Analysis (SA) is an important branch of statistics that deals with time-to-event outcomes. 
Because of its importance in many diverse areas of application, adaptation of machine and deep learning techniques to SA has received increased interest in recent years.  

Generally, we are interested in inference about the outcome of interest (time-to-event) $Y>0$, for example by estimating the distribution of the event times or their expectation. 
In SA, this is complicated by (right-, left- or interval-)censoring of individual observations as well as (left- or right-)truncation in the data.
Additional complications arise when we can observe the same event type multiple times (recurrent events) or when censoring times cannot be assumed to be independent of event times (competing risks). 

Several techniques have been developed over the years to deal with such data, prominently the non-parametric Kaplan-Meier \citep[KM;][]{kaplanNonparametricEstimationIncomplete1958} and Aalen-Johansen \citep[AJ;][]{aalenEmpiricalTransitionMatrix1978} estimators, the partial likelihood-based Cox Proportional Hazards (PH) model \citep{coxRegressionModelsLifeTables1972, coxPartialLikelihood1975}, as well as parametric likelihood-based approaches \citep{kalbfleischStatisticalAnalysisFailure2002} and their respective extensions.
Since the 2000s, machine learning (ML) approaches have been adapted to SA, including penalized Cox models \citep{simonRegularizationPathsCoxs2011,goemanL1PenalizedEstimation2010a}, SA variants of random forests \citep{ishwaranRandomSurvivalForests2008, ishwaranRandomSurvivalForests2014, jaegerObliqueRandomSurvival2019}, boosting based approaches \citep{Schmid2008, binderBoostingHighdimensionalTimeevent2009, reulen2015boostingmultistate}, and Bayesian methods \citep{Sparapani2016, madjar2021}.

The methodological adaptation of AI methods to SA has been summarized comprehensively in a survey by \cite{wang_machine_2019} for ML methods and a review by \cite{wiegrebe.deep.2024b} for Deep Learning (DL) methods. 
While this illustrates that AI-based SA is a growing field of research, various factors limit its applicability, particularly for practitioners: 

\begin{enumerate}
    \item[(a)] Both \cite{wang_machine_2019} and \cite{wiegrebe.deep.2024b} identify only few methods that explicitly go beyond the single-event, right-censored data setting.
    \item[(b)] Many research papers only provide insufficient code or proof-of-concept implementations that are hard to use in standard machine learning workflows which require standardized pre-processing, resampling, hyperparameter tuning and evaluation capabilities in addition to the actual learning algorithm.
    \item[(c)] There is often a considerable delay between the development and implementation of a method for regression or classification tasks and its adaptation to survival analysis.
\end{enumerate}

\begin{figure}[!ht]
\begin{center}
  \includegraphics[width=1\textwidth]{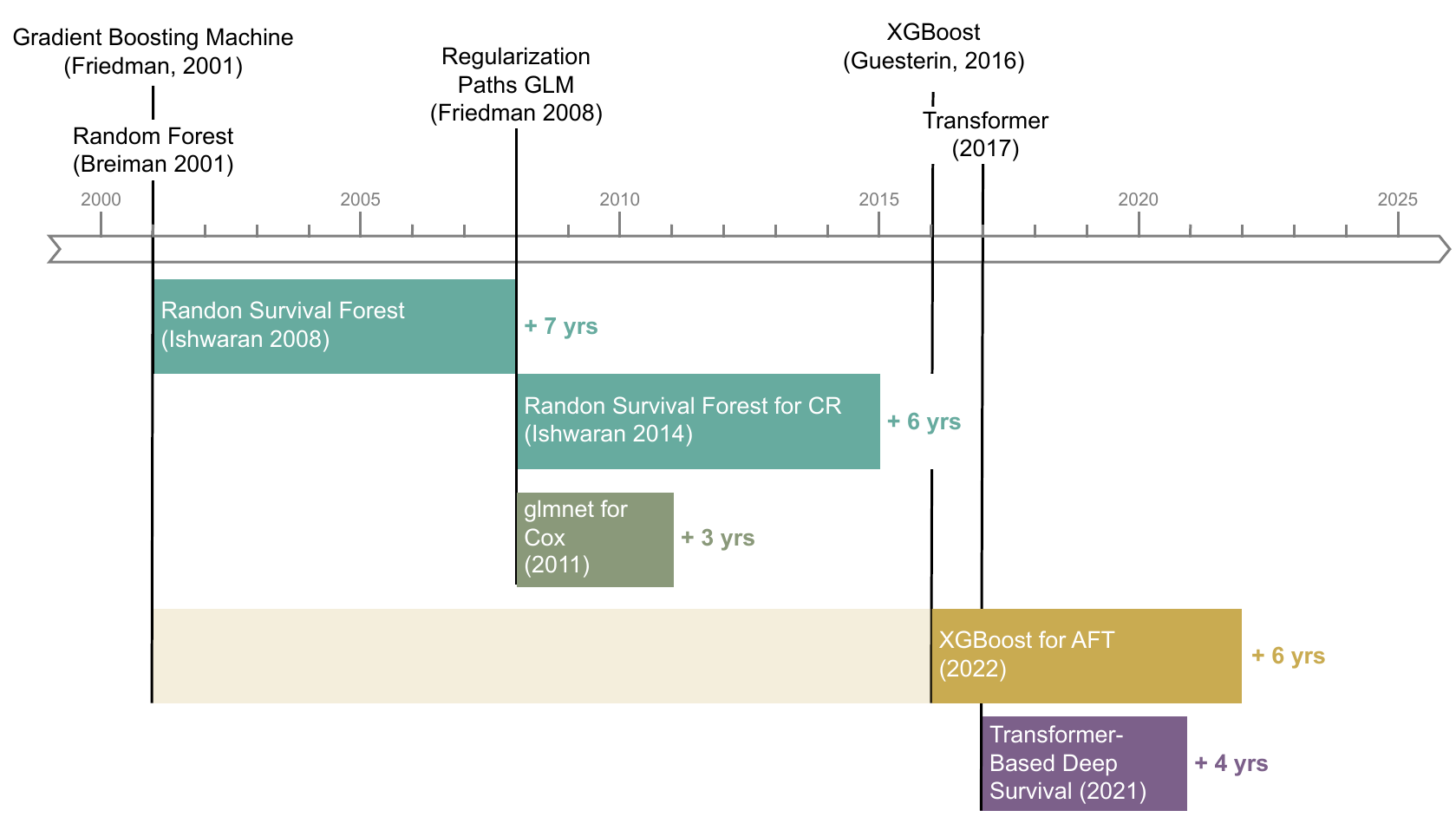}
\end{center}
\caption{Timeline illustrating the delay between development of new ML methods or software for regression and classification and their adaptation to survival analysis.}
\label{fig:ml-sa-delay}
\end{figure}

This is illustrated in Figure \ref{fig:ml-sa-delay}. 
There was a 7-year delay between the proposal of random forests for regression and classification in \cite{breiman2001random} and its adaptation to single-event, right-censored data in \cite{ishwaran_rsf_2008}. 
The extension to competing risks \citep{ishwaranRandomSurvivalForests2014} took another 6 years. 
Similar delays can be observed for boosting \citep{chen2016xgboost, barnwalXgboostSurvival2022}, regularized regression models \citep{friedmanRegularizationGLM2008, simonRegularizationPathsCoxs2011}, and deep learning \citep[e.g.,][]{transformers2017, transformerDeepSurvival2021}.

Available adaptations in research papers often suffer from point (b), while implementations in popular general purpose software often suffer from point (a). 
For example, the popular XGBoost library \citep{chen2016xgboost} that was used in many winning submissions in Kaggle competitions was extended to fit accelerated failure time models 6 years after the initial publication (although the software had been around for longer), but only learns the scale parameter of the distribution and thus does not provide the full distribution estimation. 
There is now also an implementation of the Cox model in the XGBoost software, but it natively only returns the risk-score. 
Both implementations only deal with single-event, right-censored data.

\begin{figure}[!ht]
    \centering
\includegraphics[width=1\linewidth]{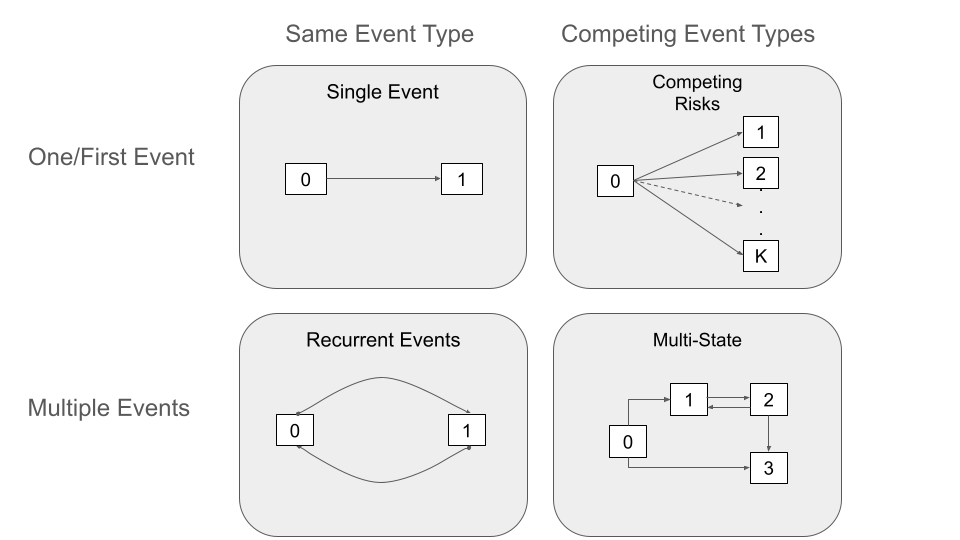}
\caption{Schematic representation of different survival tasks. Rows indicate if we observe (or are interested in) time to one or first event, or if we are interested in and have observed multiple events. Columns indicate whether we consider occurrences of a single event type or different competing event types.}
\label{fig:eha}
\end{figure}

In this work, we consider reduction techniques for survival analysis, that is, techniques that transform a survival task to a more standard regression or classification task. We categorize survival tasks into two dimensions, as these also dictate the pre- and post-processing steps as well as the applicability of the different reduction techniques.
The first dimension is the type of censoring or truncation, where we differentiate between left-, right- and interval-censoring, as well as left- and right-truncation (see \citet{wiegrebe.deep.2024b} for examples).
The second dimension is the number and type of events observed or equivalently the type of transitions we want to model. 
The latter is illustrated in Figure \ref{fig:eha}, where we differentiate between the number of events observed per observation unit (one/first or multiple) and the number of event types observed (one/same or competing event types). 
Arrows indicate possible transitions between different states, usually starting from an initial state 0. 
The most general case is the multi-state process (bottom right) with multiple transitions and the possibility of back-transitions (see Section \ref{ssec:beyond-single-event}).

Ideally, we want our ML and DL methods to be able to deal with as many of these survival tasks as possible. 
As mentioned before, however, many published methods for ML-based survival analysis focus on single-event, right-censored data. 
As we will show later, some reduction techniques cover many - though not all - of these settings, and there are variations depending on the reduction technique of choice.
These techniques

\begin{figure}[!ht]
    \begin{center}
        \includegraphics[width=1\textwidth]{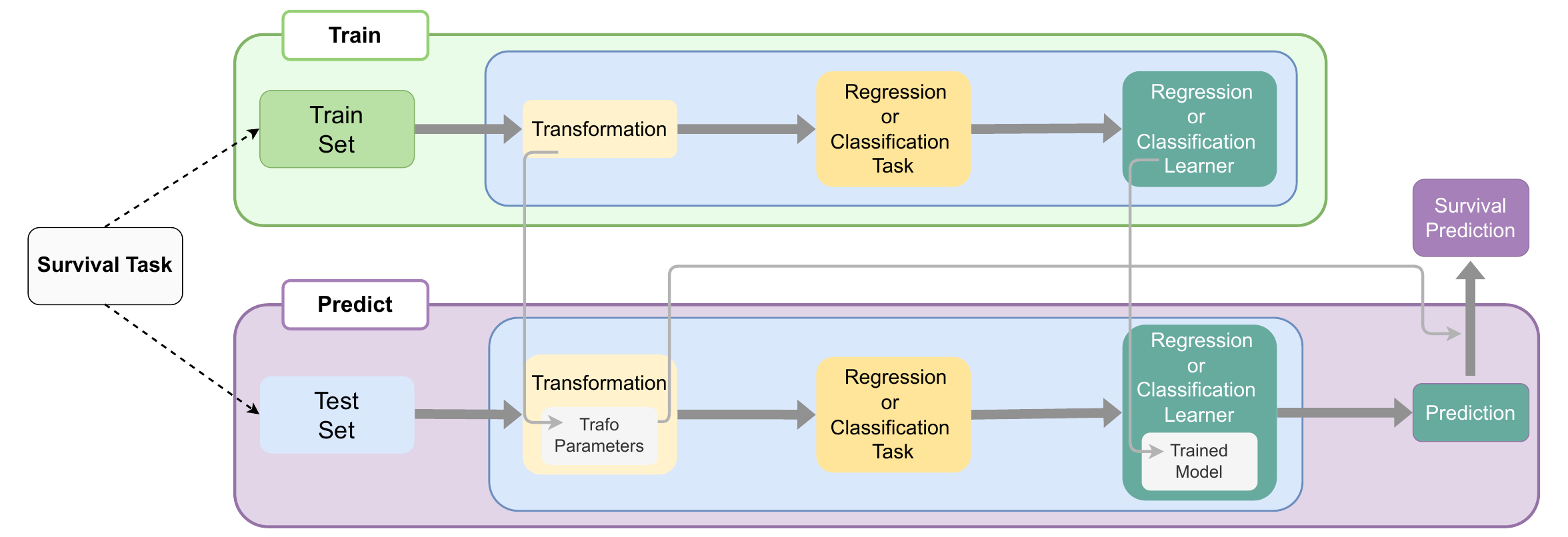}
    \end{center}
    \caption{Schematic representation of a reduction ML pipeline. The survival task is split into training and test data. In the training step (top), the survival task is transformed to a regression or classification task, depending on the chosen reduction method and specifics of the underlying survival task. A regression or classification learner is then trained on the transformed data. During the prediction step (bottom), new test data undergo the same transformation using parameters derived during training. Subsequently, the trained model generates predictions in the regression or classification domain, which are then mapped back to survival prediction types according to the reduction strategy and the transformation parameters.}
    \label{fig:reduction-pipeline}
\end{figure}

\begin{itemize}
    \item are valid methods for survival analysis that appropriately deal with censoring and/or truncation,
    \item do not make (strong) assumptions about the underlying distribution of event times,
    \item are applicable to many survival tasks, including competing risks and multi-state settings,
    \item can predict different quantities of interest in SA, including (discrete) hazards, survival probabilities, cumulative incidence functions, conditional on features,
    \item can use any off-the-shelf implementation of ML or DL methods for regression or classification (depending on the reduction technique),
    \item can (explicitly) model time-varying effects and thus deal with non-proportional hazards,
    \item and have competitive performance compared to specialized survival learners.
\end{itemize}
\ \\
The general procedure is visualized in Figure \ref{fig:reduction-pipeline}: Reduction techniques for SA transform a survival task to a regression or classification task through a dedicated pre-processing step. While the specifics of the pre-processing depend on the choice of reduction technique and the specific survival task at hand (type of censoring/truncation, presence of competing risks, etc.), upon completion the transformed data can be analyzed using standard methods without further modifications to the learner of choice or the need for SA-specific loss functions. In a predictive modeling context, the transformation parameters derived from the training data are reused to process the test data and the model trained on the transformed training set is then applied to generate predictions, which are subsequently mapped back to survival outputs (survival probabilities, cumulative incidence functions, etc.), if necessary.

The reduction techniques considered in this work are piecewise exponential models (PEM), discrete-time (DT) methods, pseudo values (PV), censoring weighted binary classification (IPCW) and continuous ranking method (CRM), which we group into reductions that estimate the entire event time distribution (PEM, DT) and reductions that are primarily used to obtain a point estimates of a quantity of interest. 
While the individual techniques that will be discussed in this work have been proposed separately in the literature as stand-alone methods for SA in their own right and partially adapted to the ML setting, to our knowledge they have not been considered jointly as reduction techniques.

The contributions of this work are: 

\begin{itemize}
    \item a unified framework for reduction technique pipelines for survival analysis
    \item overview of strengths and weaknesses of the different techniques and their applicability to different survival tasks
    \item a unified, user-friendly implementation of various reduction techniques that can be integrated into a standard machine learning workflows
    \item quantitative benchmark comparison of the different reductions compared to established statistical machine learning methods for SA
\end{itemize}

Table \ref{tab:overview-table} provides an overview of the reduction techniques discussed in this paper, along with their respective prediction type, target of estimation, target/label, and survival quantities of interest.

\begin{table}[p]
\centering
\begin{sideways}
\renewcommand{\arraystretch}{1.15}
\small
\begin{tabular}{|p{2.0cm}|p{4.5cm}|p{4.5cm}|p{1.7cm}|p{1.7cm}|p{4.9cm}|}
\hline
\textbf{} & 
\textbf{PEM} \newline (Sec. \ref{ssec:methods-pem}) &
\textbf{DT} \newline (Sec. \ref{ssec:methods-dt}) &
\textbf{IPCW} \newline (Sec. \ref{ssec:methods-ipcw}) &
\textbf{CRM} \newline (Sec. \ref{ssec:methods-crm}) &
\textbf{PV} \newline (Sec. \ref{ssec:methods-pv}) \\
\hline

\textbf{Prediction Type} &
Distribution &
Distribution &
Point \newline Estimate &
Point \newline Estimate &
Point \newline Estimate \\
\hline

\textbf{Target of Estimation} &
$h(\tau|\mathbf{x})$ \newline (Eq. \ref{eq:pem-hazard}) &
$h(j_{\tau}|\mathbf{x})$ \newline (Eq. \ref{eq:discrete_hazard}) &
$\pi(\mathbf{x})$ \newline (Eq. \ref{eq:IPCW-pred-target}) &
$r(\mathbf{x})$ \newline (Eq. \ref{eq:crm-pred-target}) &
$\theta(\tau|\mathbf{x})$ \\
\hline

\textbf{Target/ Label} &
$d_{ij}$ \newline (Eq. \ref{eq:partition-event-indicator}) &
$d_{ij}$ \newline (Eq. \ref{eq:partition-event-indicator}) &
$e_i$ \newline (Eq. \ref{eq:IPCW-event-label}) &
$r_i$ \newline (Eq. \ref{eq:crm-pred-label}) &
$\theta_i$ \newline (Eq. \ref{eq:pv-def}) \\
\hline

\textbf{Prediction Task} &
Regression &
Classification &
Classification &
Regression &
Regression \\
\hline

$\hat{S}(\tau|\mathbf{x})$ &
$\exp(-\sum_j^{j_{\tau}}$ $\hat{h}(t_j|\mathbf{x}) \cdot t_j)$ \newline (Eq. \ref{eq:hazard-to-surv}) &
$\prod_{l=1}^{j_{\tau}} (1-\hat{h}(l|\mathbf{x}))$ \newline (Eq. \ref{eq:discrete_survival_function}) &
$1-\hat{\pi}(\mathbf{x})$ &
- &
${\widehat{\theta}}_i(\tau) = n{\widehat{S}(\tau)} - (n-1){\widehat{S}(\tau)}^{-i}$\newline $\widehat{S}(\tau)$ estimated from the Kaplan-Meier estimator
\newline (Eq. \ref{eq:POdef}) \\
\hline

$\widehat{CIF}_k(\tau|\mathbf{x})$ &
$\int^\tau_0 h_k(u) S(u) \ du$ \newline (Eq. \ref{eq:cif}) &
$\sum_{l=1}^{j_{\tau}} \hat{h}_k(l|\mathbf{x}) \cdot$ $\hat{S}(l-1|\mathbf{x})$ \newline (Eq. \ref{eq:discrete_cif}) &
- &
- &
 $\widehat{\theta}_{ki}(\tau) = n \widehat{CIF_k}(\tau) - (n - 1) \widehat{CIF_k}^{-i}(\tau)$\newline 
 $\widehat{CIF_k}(\tau)$ estimated from the Aalen-Johansen estimator\newline 
 (Eq. \ref{eq:pv:MSMdef-transprob}) \\
\hline

$\hat{\mathbf{P}}(s, \tau)$ &
$\Prodi_{u \in [s, \tau)}$ $\left( \textbf{I} + \textbf{d}\hat{\textbf{H}}(u|\mathbf{x})\right)$ \newline (Eq. \ref{eq:trans-prob}) &
$\prod\limits_{j=j_s}^{\,j_\tau}$
$\left( \textbf{I} + \textbf{d}\hat{\textbf{H}}(j|\mathbf{x})\right)$ \newline
with $\hat{\textbf{H}}(j|\mathbf{x}) := \left( \hat{h}_{o,l}(j|\mathbf{x}) \right)_{o,l}$ 
and $\hat{h}_{o,o} = 1 - \sum\limits_{l \neq o}\hat{h}_{o,l}(j|\mathbf{x})$ &
- &
- &
 $\widehat{\theta}_{ki}(\tau) = n \widehat{P_k}(\tau) - (n - 1) \widehat{P_k}^{-i}(\tau)$\newline 
 $\widehat{P_k}(\tau)$ estimated from the Aalen-Johansen estimator\newline 
 (Eq. \ref{eq:pv:MSMdef-transprob} \\
\hline

\textbf{RMST} &
$\int_0^{\tau} \hat{S}(u|\mathbf{x})\mathrm{d}u$ \newline (Eq. \ref{eq:rmst}) &
$\sum_{l=1}^{j_{\tau}} \hat{S}(l|\mathbf{x})$ \newline (Eq. \ref{eq:rmst}) &
- &
- &
${\widehat{\theta}_{i}(\tau) = n\int_0^{\tau} {\widehat{S}(t)}dt - (n-1)\int_0^{\tau} {\widehat{S}(t)}^{-i}dt}$
\newline $\widehat{S}(\tau)$ estimated from the Kaplan-Meier estimator
\newline (Eq. \ref{eq:PORMSTdef}) \\
\hline

\end{tabular}
\end{sideways}
\caption{Overview of five reduction techniques for survival analysis in terms of their prediction type, target of estimation, target/label, prediction task, and survival quantities of interest.}
\label{tab:overview-table}
\end{table}


In Section \ref{sec:notation} we introduce notation and definitions relevant for the remainder of this work. 
Sections \ref{sec:partition-based-reductions} and \ref{sec:timepoint-based-reductions} introduce the different reduction techniques, starting with the general definition, followed by details regarding the applicability to different survival tasks as well as limitations, and concluding with a "further reading" section.
Section \ref{sec:software} discusses software implementations for reduction techniques, particularly in a ML context.
In Section \ref{sec:applications}, we illustrate the application of the respective reduction techniques to selected data examples, followed by a quantitative comparison to established statistical and ML methods for SA in Section \ref{sec:benchmarks}.
Section \ref{sec:discussion} summarizes results and limitations and provides ideas for future avenues of research.

\section{Notation and definitions}
\label{sec:notation}
In the following we introduce common quantities that are usually targets of estimation or prediction in survival analysis. For simplicity, these terms are introduced without dependence on features, but the equations are equally valid in presence of features.

\subsection{Quantities of interest}
\label{ssec:quantities}

In this section, we briefly recap different functions used to represent (summaries of) the event time distribution, frequently targets of estimation in SA. Let $Y>0$ be the random variable representing event times with realizations $y$ and let $\tau > 0$ be some arbitrary time point.
The hazard function, given by
\begin{equation}
\label{eq:chazard}
h(\tau) = \lim_{\Delta\searrow 0}\frac{P(\tau < Y \leq \tau + \Delta|Y \geq \tau)}{\Delta},
\end{equation}
is the instantaneous risk to observe an event given that the event has not been observed up until that point. It is a frequent target of estimation, especially among non- and semi-parametric methods such as Cox-type models.
Additionally, the hazard rate is often used to construct the survival function via the relationship
\begin{equation}
\label{eq:hazard-to-surv}
S(\tau) = P(Y > \tau) =  \exp(-H(\tau)) = \exp\left(-\int_0^\tau h(u)\ du\right),
\end{equation}
where $H_Y(\tau)$ is the cumulative hazard function.

Finally, we define the restricted mean survival time (RMST) \citep{irwin1949pseudo, zhao2016} as 
\begin{equation}
\label{eq:rmst} 
\mu_{\tau} = \mathbb{E}(\min (Y, \tau)) = \int_0^{\tau} S(u)\mathrm{d}u, 
\end{equation}
which has become popular recently, especially as an intuitive and clinically meaningful measure of feature and treatment effects in randomized clinical trials, in particular when the PH assumption is violated \citep{Royston2011, Royston2013}.



\subsection{Beyond single-event setting}
\label{ssec:beyond-single-event}
As depicted in Figure \ref{fig:eha}, a time-to-event process can be represented in terms of transitions between different states. States can be \emph{transient} (transitions to another state possible) or \emph{absorbing} (no further transitions possible, e.g., death). 

In the competing risks setting, some states may be transient, but the focus is on the first of multiple mutually exclusive events, without modeling further transitions. Thus, there are $q$ distinct competing states, allowing transitions of the form $0 \rightarrow k$, with $k \in \{1, \dots, q\}$. 

Recurrent events describe a setting with repeated occurrences of the same non-terminal event, such as non-fatal respiratory infections. If, in addition, a terminal event is present, it needs to be accounted for, usually by casting it as a multi-state problem. 

Single-event, recurrent events, and competing risks settings can be viewed as special cases of the multi-state setting. Let $K \in \{1,\ldots,q\}$ be a random variable representing a state (competing risks) or transition (e.g., $0\rightarrow1$, $0\rightarrow3$, ..., $2\rightarrow3$ in Figure \ref{fig:eha}).
Finally, let $k$ denote realizations of $K$.
We extend Equation \eqref{eq:chazard} to the general multi-state transition rate
\begin{equation}
\label{eq:transition-hazard}
    h_{k}(\tau) = \lim_{\Delta \searrow 0}\frac{P(\tau \leq Y \leq \tau + \Delta, K = k|Y \geq \tau)}{\Delta}.
\end{equation}
In the competing risks setting, $h_k$ are the cause-specific hazards $h_{0\rightarrow k}$. In the multi-state setting, these are transition-specific hazards $h_{o\rightarrow \ell}(\tau)$, where $o$,$\ell$ are the initial state and end state, respectively.

The all-cause hazard, which is the total hazard of all possible transitions, is defined as
\begin{equation}
\label{eq:marginal-transition-hazard}
    h(\tau) = \sum_{k=1}^{q}h_{k}(\tau).
\end{equation}

Once the transition hazards for the different settings have been estimated, further quantities of interest can be obtained from them using closed formulae. 
In the single-event case, this is typically the survival curve (Equation \eqref{eq:hazard-to-surv}). 
In the competing risks setting, we are often interested in the cumulative incidence function (CIF)

\begin{equation}
    \label{eq:cif}
    CIF_k(\tau) = P(Y \leq \tau, K = k) = \int^\tau_0 h_k(u) S(u) \ du,
\end{equation}
where $S(u)$ is the all-cause survival probability derived via Equations \eqref{eq:marginal-transition-hazard} and \eqref{eq:hazard-to-surv}.

In the multi-state setting, we are often interested in the transition probabilities $P_{o\ell}(s, \tau)$, i.e., the probability to transition from state $o$ to state $\ell$ between times $s$ and $\tau$. These can be obtained via the empirical transition matrix \citep{aalenEmpiricalTransitionMatrix1978}

\begin{equation}
    \label{eq:trans-prob}
    \mathbf{P}(s, \tau) = \Prodi_{u \in [s, \tau)} \left( \textbf{I} + d\textbf{H}(u)\right),
\end{equation}
where $d\mathbf{H}(\tau)$ is the matrix of cumulative hazard rate differences $H_{k}(\tau + \Delta\tau) - H_{k}(\tau)$ (cf. \citet{beyersmannCompetingRisksMultistate2012}), which can be calculated directly from estimates of Equation \eqref{eq:transition-hazard}.



\section{Reductions for partition-based hazard estimation}
\label{sec:partition-based-reductions}
\subsection{Partitioning of the follow-up}
\label{ssec:partitioning}
\begin{figure}[!ht]
    \centering
    \includegraphics[width=0.8\linewidth]{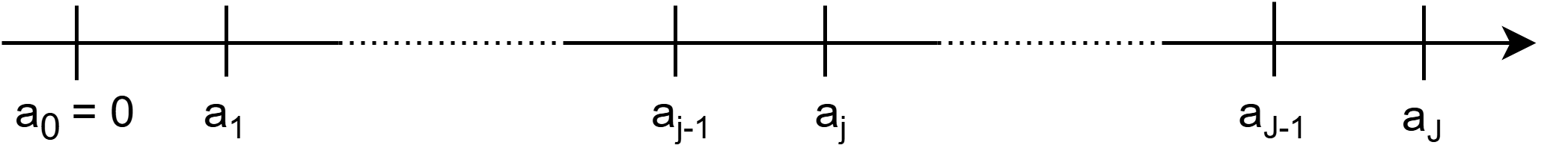}
    \vspace{-0.4cm}
    \caption{Schematic representation of the partitioning of continuous follow-up time using cut points $a_j$, $j \in \{1, \dots, J\}$, resulting in $J$ (here equidistant) time intervals. The corresponding partitioning of the survival dataset is illustrated in Table \ref{tab:data-partitioning}.}
    \label{fig:fu-partitioning}
\end{figure}

Some reduction techniques partition the time axis into $J$ intervals, estimating a constant or discrete-time hazard within each. As shown in Figure \ref{fig:fu-partitioning}, the $j$-th interval is $I_j := (a_{j-1}, a_j]$, and $J_i \in {1, \dots, J}$ denotes the interval containing the observed time of subject $i$, $t_i$, with $t_i \in I_{J_i} = (a_{J_i-1}, a_{J_i}]$ and $i \in 1,\dots,n$. Intervals may be equidistant or vary in length.

This partitioning is used to create a new long-form dataset with one row for each subject $i$ and each time interval for which $i$ was still (at least partially) at risk (i.e., $\forall j \in \{1, \dots, J_i\}$). This is done by defining the following variables for each time interval $I_j$ and subject $i$ (with corresponding observed time $t_i$, which is the minimum of the subject's true event time $y_i$ and their censoring time $c_i$):

\begin{itemize}
    \item \refstepcounter{equation}
          the event-specific event indicator
          \begin{align}
            d_{ij} = \begin{cases} 
                1 & \text{if } t_i \in I_j \wedge d_i = 1 \\
                0 & \text{else}
            \end{cases} \tag{\theequation}\label{eq:partition-event-indicator},
          \end{align}
    \item the time at risk 
        \begin{align*}
            t_{ij} = \begin{cases} a_j - a_{j-1} & \text{if } a_j < t_i\\ t_i - a_{j-1} & \text{if } a_{j-1} < t_i \leq a_j\end{cases},
        \end{align*}
    \item and an "offset" to model a rate of events per unit time
        \begin{align*}
            o_{ij} = \log (t_{ij}),
        \end{align*}
\end{itemize}

where $d_i$ is the usual status indicator.

This data transformation is illustrated in Table \ref{tab:data-partitioning}.
Left-truncation, time-varying features, and time-varying effects all require similar data transformations and can thus be naturally incorporated. 
In particular, since $t_{ij}$ becomes just another feature in this transformed data, time-varying effects simply become interaction terms with time. Additional cut points may be necessary, e.g., at the times when time-varying features change.

\begin{table}[!ht]
\centering
\begin{tabular}{c c}
\hspace{-0.9cm}
\begin{minipage}{0.18\textwidth}
\vspace{-0.5cm}
\centering
\begin{tabular}{|c|c|c|c|}
\hline
\textbf{i} & \textbf{t}$_i$ & $\textbf{d}_i$ & \textbf{age}$_i$ \\ \hline
1 & 1.3 & 1 & 31 \\ \hline
2 & 0.5 & 0 & 67 \\ \hline
3 & 2.1 & 1 & 42 \\ \hline
\end{tabular}
\end{minipage}
&
\hspace{-0.4cm}
\begin{minipage}{0.25\textwidth}
\centering
\scriptsize 
\vspace{0.5cm} 
no left-truncation \par
\vspace{0.2cm} 
\scalebox{4}{$\nearrow$} \\[1cm]
\scalebox{4}{$\searrow$} \par
\vspace{0.2cm} 
left-truncation of \par
subject 1 at 0.5 \par
and of subject 3 at 1.5
\vspace{0.5cm} 
\end{minipage}
\hspace{-0.4cm}
\begin{minipage}{0.5\textwidth}
\centering
\begin{tabular}{|c|c|p{1.0cm}|c|c|c|c|c|}
\hline
\textbf{i} & \textbf{j} & \centering \textbf{I}$_j$ & $\textbf{d}_{ij}$ & \textbf{t}$_{ij}$ & \textbf{o}$_{ij}$ & \textbf{a}$_j$ & \textbf{age}$_i$ \\ \hline
1 & 1 & (0,0.5]   & 0 & 0.5 & -0.7   & 0.5   & 31 \\ \hline
1 & 2 & (0.5,1]   & 0 & 0.5 & -0.7   & 1     & 31 \\ \hline
1 & 3 & (1,1.5]   & 1 & 0.3 & -1.2   & 1.5   & 31 \\ \hline
2 & 1 & (0,0.5]   & 0 & 0.5 & -0.7   & 0.5   & 67 \\ \hline
3 & 1 & (0,0.5]   & 0 & 0.5 & -0.7   & 0.5   & 42 \\ \hline
3 & 2 & (0.5,1]   & 0 & 0.5 & -0.7   & 1     & 42 \\ \hline
3 & 3 & (1,1.5]   & 0 & 0.5 & -0.7   & 1.5   & 42 \\ \hline
3 & 4 & (1.5,2]   & 0 & 0.5 & -0.7   & 2     & 42 \\ \hline
3 & 5 & (2,2.5]   & 1 & 0.1 & 0      & 2.5   & 42 \\ \hline
\end{tabular}

\vspace{0.5cm}

\begin{tabular}{|c|c|p{1.0cm}|c|c|c|c|c|}
\hline
\textbf{i} & \textbf{j} & \centering \textbf{I}$_j$ & $\textbf{d}_{ij}$ & \textbf{t}$_{ij}$ & \textbf{o}$_{ij}$ & \textbf{a}$_j$ & \textbf{age}$_i$ \\ \hline
1 & 2 & (0.5,1]   & 0 & 0.5 & -0.7   & 1     & 31 \\ \hline
1 & 3 & (1,1.5]   & 1 & 0.3 & -1.2   & 1.5   & 31 \\ \hline
2 & 1 & (0,0.5]   & 0 & 0.5 & -0.7   & 0.5   & 67 \\ \hline
3 & 4 & (1.5,2]   & 0 & 0.5 & -0.7   & 2     & 42 \\ \hline
3 & 5 & (2,2.5]   & 1 & 0.1 & 0      & 2.5   & 42 \\ \hline
\end{tabular}
\end{minipage}
\end{tabular}
\caption{Illustration of survival data transformation due to partitioning of follow-up time. Starting from a standard survival dataset with a single row per subject $i$ containing the ID, observed time, event indicator, and features (here: baseline age), the follow-up time is partitioned into time intervals (here: equidistant with $a_0=0$, $a_1=0.5$, $a_2=1$, etc.) as illustrated in Figure \ref{fig:fu-partitioning}. Correspondingly, the dataset is transformed into an expanded long-format dataset with one row per subject and time interval. The bottom left table additionally incorporates left-truncation ($t_1^L=0.5$ and $t_3^L=1.5$) by removing (or restricting) intervals prior to the respective subjects' left-truncation times.}
\label{tab:data-partitioning}
\end{table}

For competing risks, multi-state, and recurrent events analyses, the data must be further transformed -- or, to be precise, augmented.

In the context of competing risks, a separate dataset for each of the $q$ competing events is created, where for cause $k$ we replace the interval-specific event indicator $d_{ij}$ by a cause-specific indicator
$d_{ijk} = \begin{cases} 1 & \text{if } d_{ij} = 1 \wedge \ d_{ik} = 1\\ 0 & \mbox{else} \end{cases}$,
with $d_{ik}$ a binary indicator for whether $i$'s event was of type $k$ (or not).
In addition, we add a column \texttt{cause} to each of the $q$ datasets, taking on the value $k$ for all rows of the $k$-th dataset.
Note that in the $k$-th dataset, all events $\tilde{k} \neq k$ are implicitly treated as censoring.
This data transformation step can thus also be viewed as augmentation by counterfactual transitions: the row of a subject with event type $k$ in the $\tilde{k}$-th dataset corresponds to the counterfactual transition from state $0$ to state $\tilde{k}$ (and thus $d_{ij\tilde{k}}=0$ even in interval $j_i$). 

In multi-state processes, each state and its transitions can be viewed as nested competing risks. Like in competing risks transformations, multi-state data are stacked across all $q$ transitions, with an added categorical variable indicating the transition $k \in {0, 1, ..., q}$ taken by subject $i$. The binary event indicator $d_{ij}$, defined as in the single-event setting, indicates whether subject $i$ experiences the specific transition. Since subjects enter transition-specific risk sets at different times, the transformation includes both the transition time $t_{ij}$ and the state entry time in order to handle the resulting left-truncation. To address censoring from competing events, counterfactual transitions are added as additional rows with the same times, transition identifier, and $d_{ij\tilde{k}}=0$ $\forall \tilde{k} \neq k$.

This follow-up time partitioning and data transformation applies identically to both piecewise exponential and discrete-time reduction techniques (see Sections \ref{ssec:methods-pem} and \ref{ssec:methods-dt} below). The key difference is that piecewise exponential models retain exact times at risk $t_{ij}$ (via offsets $o_{ij}$), while discrete-time models ignore exact times at risk, relying solely on the interval index $j$. 
Importantly, although each subject now has multiple rows in the dataset, which needs to be accounted for during resampling, they can be treated as independent during model training (see details in subsequent sections), such that ML algorithms can be used without modification.


\subsection{Piece-wise exponential reduction}
\label{ssec:methods-pem}

The piece-wise exponential model (PEM) is a reduction technique that transforms a survival task to a Poisson regression task. 
Initially proposed in the early 1980s \citep{friedman1982piecewise}, it saw limited adoption compared to the Cox model \citep{coxRegressionModelsLifeTables1972, coxPartialLikelihood1975}, despite their demonstrated equivalence in some situations \citep{whiteheadFittingCoxsRegression1980}.
However, this model class has regained popularity more recently (see Section \ref{sssec:further-reading-pem}), with modern versions leveraging efficient algorithms and refined methods for the estimation of the (baseline) hazard and time-varying effects.

\subsubsection{Single-event setting}
\label{sssec:single-event-pem}
Like the Cox model, the PEM estimates the hazard \eqref{eq:chazard} as an exponential function depending on feature vector $\mathbf{x}$, i.e., 
\begin{equation}
    \label{eq:pem-hazard}
    h(\tau|\mathbf{x}_i) = \exp(g(\tau, \mathbf{x}_i)),
\end{equation}
where $g(\tau, \mathbf{x})$ is some function of the features and time, learned from the data. Under the PH assumption, we obtain the familiar Cox PH-type hazard $h(\tau|\mathbf{x}) = h_0(\tau) \cdot$ $\exp(g(\mathbf{x})) = \exp(\log(h_0(\tau)) + g(\mathbf{x}))$. In contrast to the Cox model, however, the baseline hazard is estimated jointly with the feature effects.

For the estimation, the data is transformed as illustrated in Table \ref{tab:data-partitioning} of Section \ref{ssec:partitioning}. 
The newly created status variable $d_{ij}\sim Po(\mu_{ij})$ is then used as outcome in a Poisson model, in which the time at risk $t_{ij}$ enters logarithmically as offset $o_{ij}$. 
Intuitively, the model estimates the expected conditional event rate $\mu_{ij}=h_{ij}t_{ij}$ as the product of the hazard rate of subject $i$ in the $j$th interval ($h_{ij}$) multiplied with the time subject $i$ was at risk for the event ($t_{ij}$). 
For predictions, the offset is ignored, such that the model returns $\hat{h}(\tau|\mathbf{x})$. 
Given the piece-wise constant nature of the hazard in this model, calculation of the quantities of interest in Equation \eqref{eq:hazard-to-surv} becomes straightforward.

\subsubsection{Multi-state setting}
\label{sssec:multi-state-pem}
In the multi-state setting, the aim is to estimate transition hazards (see Equation \eqref{eq:transition-hazard}). Keeping track of different transitions, the data is often given in "start-stop" notation, where each row represents one transition, given by quadruples  
\begin{align}
\label{eq:ms-tupel}
(t^{entry}_{ike}, t^{exit}_{ike}, d_{ike}, \mathbf{x}_{i}),
\nonumber
\end{align}
where $t^{entry}_{ike}$ is the observed entry (i.e., left-truncation) time of subject $i$ into the risk set for transition $k=1,\ldots,q$ in episode $e=1,\ldots,m$.
$t^{exit}_{ike}$ is the respective observed exit time, due to either the transition occurring or censoring. 
$d_{ike}$ is the corresponding status indicator and $\mathbf{x}_{i}$ the feature row vector.

\begin{table}[!ht]

\begin{center}
    \begin{tabular}{|c|c|c|c|c|c|c|}
    \hline
    \textbf{i} & \textbf{o} & \textbf{$\ell$} & \textbf{e} & \textbf{t}$^{entry}_{ike}$ & \textbf{t}$^{exit}_{ike}$ & \textbf{d}$_{ike}$ \\ \hline
    1 & 1 & 2 & 1   & 0   & 0.5 & 1   \\ \hline
    1 & 2 & 1 & 1   & 0.5 & 1   & 1   \\ \hline
    1 & 1 & 2 & 2   & 1   & 3   & 1   \\ \hline
    2 & 0 & 1 & 1   & 0   & 3   & 0   \\ \hline
    3 & 1 & 2 & 1   & 0   & 1   & 1   \\ \hline
    3 & 2 & 3 & 1   & 1   & 2.5 & 1   \\ \hline
    \end{tabular}
\end{center}
    \caption{Example data (raw, untransformed) for the multi-state model depicted in Figure \ref{fig:eha} in start-stop notation including three subjects, transitions, episodes, and the feature age. The five possible transitions $o \rightarrow \ell$ are $0 \rightarrow 1$, $1 \rightarrow 2$, $2 \rightarrow 1$, $0 \rightarrow 3$, and $2 \rightarrow 3$.}
    \label{tab:msm-rawdata}
\end{table}

With this data structure at hand, we estimate the transition hazard as
\begin{equation}
    \label{eq:pem-hazard-msm}
    h_{ke}(\tau|\mathbf{x}_i) = \exp(g(\tau, k, e, \mathbf{x}_i)).
\end{equation}
The multi-state framework includes the competing risks setting (dropping $e$) and the recurrent events setting (dropping $k$) as special cases.

The advantage of the PEM approach is that we can use the same estimation in multi-state models as for single-events; only the data transformation differs. Unlike in the single-event transformation, the multi-state version must store information on state entry (because multiple transitions can happen with progressing time) and exit time. It must also include information on possible exit states for each state. Other competing events are considered censoring for the event of interest, i.e., $d_{ike} = 0$, see $i = 3$ in Table \ref{tab:msm-rawdata}.
Essentially, we either create one dataset for each possible transition (and episode) or include $k$, $e$ as features in our hazards.
The status indicator is then given as 
\begin{equation}
    d_{ijke} = I(d_{ike} = 1 \wedge t_i \in I_j) \in \{0,1\} \sim Po(\mu_{ijke})
\end{equation}
and again used as outcome in a Poisson model, in which the time at risk $t_{ij}$ enters logarithmically as offset $o_{ij}$. Hence, the optimization of the model remains identical to the single-event setting. The hazard can be denoted as  
\begin{align}
h_{ke}(\tau|\mathbf{x})
  & = \exp (\beta_{0ke} + f_{0ke}(\tau) + g(\tau, k, e, \mathbf{x})),
\label{eq:msm-transition-hazard}
\end{align}
where the baseline hazard is given by $h_{0ke}(\tau) = \exp(\beta_{0ke} + f_{0ke}(\tau))$, including baseline effects and (non-linear) interactions of transitions $k$ and episodes $e$ with time $\tau$. The function $g(\tau, k, e, \mathbf{x})$ is again some function of the features and time, learned from the data, and can also vary across different transition trajectories and episodes. 

Given the piece-wise constant nature of the hazard in this model, the calculation of the quantities of interest in \eqref{eq:trans-prob}, especially the construction of $dH_{ke} = H_{ke}(\tau + \Delta \tau | \mathbf{x}) - H_{ke}(\tau | \mathbf{x})$, becomes straight forward. In the competing risks setting, given the hazard $h_k(\tau|\mathbf{x})$, the calculation of \eqref{eq:cif} simplifies.

\subsubsection{Limitations}
\label{sssec:pem-limitations}
In general, PEMs are limited when it comes to handling left- and interval-censored data, but they can accommodate left-truncated and right-censored data through partitioning of the follow-up. 
This transformation expands the dataset, which can slow down estimation. 
However, by reducing the problem to a Poisson regression, discretizing follow-up time preserves the essential information while also serving as a tuning parameter that balances speed and precision.
Alternatively, techniques for efficient estimation of Poisson regression in the big data context are applicable \citep{wood_gigadata_2017, reulen2015boostingmultistate, sennhenn-reulen2016structuredfusion}. Whenever hazards are not the quantity of interest -- e.g., CIF in competing risks settings, transition probabilities in multi-state settings, or RMST for treatment comparison -- post-processing the results of PEM-based survival analysis requires some effort. Software packages, e.g., \texttt{pammtools}, already introduce convenience functions for many important quantities.

\subsubsection{Further reading}
\label{sssec:further-reading-pem}
Focusing on the limitations of partitioning while still using Poisson Generalized Additive Models to estimate hazards, \citet{argyropoulosAnalysisTimeEvent2015} presents the Gauss-Labatto quadrature rule (GL) to reduce the needed number of nodes in the partitioned data. 
Piecewise additive mixed models \citep[PAMMs;][]{bender.generalized.2018} address the arbitrary choice of cut-points by using a fine grid and semiparametric, penalized estimation to model the baseline hazard and time-varying effects, enabling flexible modeling of complex time-dependent features—including weighted cumulative exposures, distributed lag non-linear effects, event-specific hazards, and feature effects.
Exploiting this flexibility, \citet{ramjithRecurrentEventsAnalysis2022} discuss the applicability and capability of analyzing recurrent events data with PAMMs, which allows the modeler to include frailties as random effects in the model. The article concludes that PAMMs offer a powerful alternative to Cox-based models, shared frailty models, and variance-adjustment methods for recurrent events, due to their ability to flexibly model complex dependencies and their practical implementation tools.
Extending the estimation flexibility following the piecewise exponential reduction, \citet{bender_general_2021} implemented and evaluated a gradient boosted trees (GBT) algorithm, using the XGBoost library. The authors show that GBT matches the strong performance of methods like ORSF and DeepHit across diverse datasets, while requiring less development effort, and note that using sparser cut-points to reduce long-form data size does not impact model performance.

Due to PEM reduction, classical variable selection schemes are available. In more complex settings with a variety of different transitions,  e.g., multi-state models, \citet{reulen2015boostingmultistate} use boosting to perform a data-driven variable selection. Boosting explores information about conditional transition-type-specific hazard rate functions by estimating the influencing effects of explanatory variables. 
Alternatively, \citet{sennhenn-reulen2016structuredfusion} extend the LASSO to structured fusion LASSO tailored for multi-state models. The extension uses $L1$-penalty and a structured fusion approach, with the focus being on relationships between different transitions for multi-state models.

For an illness-death model, a special case of a three-state multi-state model, \citet{cottin2022idnetwork} suggest IDnetwork, a deep learning architecture for disease prognostication. Inspired by \citet{lee2018deephit}, the architecture combines a shared subnetwork with three transition-specific subnetworks using multi-task learning with hard parameter sharing to capture common and unique patient patterns. Compared to multi-state Cox models (including spline variants), IDnetwork improves discrimination and calibration, particularly with non-linear data patterns in simulated and real-world cancer datasets.

\subsection{Discrete-time reduction}
\label{ssec:methods-dt}

\noindent
Discrete-time (DT) methods are particularly useful when event times are \emph{intrinsically} discrete \citep{tutz2016modeling}. However, here we consider them as an approximation for continuous time distributions. To this end, the time-axis can be partitioned into non-overlapping time intervals (as for PEMs, see Section \ref{ssec:partitioning}). 
Analogously to PEMs reducing survival tasks to Poisson regression tasks, discrete-time models reduce survival tasks to binary classification tasks and are therefore very popular among DL-based survival analysis methods \citep{wiegrebe.deep.2024b}.

Throughout this section, let $\tilde{Y} \in \{1,\ldots,J\}$ be the discrete or discretized time. In the latter case, we have $P(Y \in I_j) \Leftrightarrow P(\tilde{Y} = j)$. 

The DT hazard
\begin{equation}
    h(j|\mathbf{x}) = P(\tilde{Y} = j | \tilde{Y} \geq j, \mathbf{x})
\label{eq:discrete_hazard}
\end{equation}
represents the probability of the event occurring during the $j$-th time interval, conditional on surviving up until the beginning of that interval (see \citep{tutz2016modeling} for details). 

As for the PEM, once the hazard is estimated by a binary classification model (which returns probabilities), other quantities can be directly calculated from the hazard. 

For example, the DT survival function is given by
\begin{equation}
    S(j|\mathbf{x}) := P(\tilde{Y}>j|\mathbf{x}) = P(Y > a_j|\mathbf{x}) = \prod_{l=1}^j (1-h(l|\mathbf{x})).
\label{eq:discrete_survival_function}
\end{equation}
The unconditional probability of the event occurring at time $j$ is
\begin{equation}
    P(\tilde{Y} = j|\mathbf{x}) = h(j|\mathbf{x}) S(j-1|\mathbf{x}).
\label{eq:discrete_event_probability}
\end{equation}
Ignoring the exact event time within an interval thus gives likelihood contributions $P(\tilde{Y}=j|\mathbf{x})$ for subjects that experience an event in interval $j$ and $P(\tilde{Y} > j|\mathbf{x})$ for subjects censored in interval $j$. 
With this, and using the binary event indicators $d_{ij}$ from Section \ref{ssec:partitioning}, the likelihood for the observed data can thus be written as (cf. \citet{tutz2016modeling}): 
\begin{align}
L &\propto \prod_{i=1}^n P(\tilde{Y}_i=J_i|\mathbf{x}_i)^{d_i} P(\tilde{Y}_i > J_i|\mathbf{x}_i)^{1-d_i} \nonumber \\
&= \prod_{i=1}^n \prod_{j=1}^{J_i} h(j|\mathbf{x}_i)^{d_{ij}} 
(1-h(j|\mathbf{x}_i))^{1-d_{ij}}. 
\end{align}
This likelihood, however, is equivalent to the likelihood of binary responses $d_{ij}$ (with $i=1, \ldots, n$ and $j=1, \ldots, J_i$) from a binary response model $g(P(d_{ij}=1|\mathbf{x}_i))=f(\mathbf{x}_{i})$ with link function $g()$, some predictor function $f()$ and $d_{ij} \stackrel{iid}{\sim} Ber(h(j|\mathbf{x}_i))$.

\subsubsection{Single-event setting}
\label{sssec:single-event-dt}
In the single-event setting, any classification algorithm that returns (calibrated) probabilities can be applied to the transformed data.
DT survival approaches typically parametrize the discrete hazard function.
Thus a simple DT model is the \emph{logistic model} (or \emph{continuation ratio model}), a standard logistic regression model (i.e., using a logit link) for $h(j|\mathbf{x}_i)$, the conditional probability of the event taking place in $j$:
\begin{equation}
h(j|\mathbf{x}_i)=P\left(d_{ij}=1 | \mathbf{x}_{i}\right)=\frac{\exp \left(\eta_{ij}\right)}{1+\exp \left(\eta_{ij}\right)},
\label{eq:logistic_model_hazard}
\end{equation}
with linear predictor $\eta_{ij} = \beta_{0j}+\mathbf{x}_{i}^{\top} \boldsymbol{\beta}$ and discrete baseline hazard $\beta_{0j}$. 

However, $h(j|\mathbf{x}_i)$ can also be estimated using random forests or gradient boosting for classification (see Sections \ref{sec:applications} and \ref{sec:benchmarks} for application examples).
Section \ref{sssec:further-reading-dt} provides an overview of specific DT methods proposed in the literature.

\subsubsection{Competing risks setting}
\label{sssec:competing-risk-dt}

We now consider the CR setting introduced in Section \ref{ssec:beyond-single-event}.
In the discrete-time context, the cause-specific hazard is defined as 
\begin{equation}
    h_k(j|\mathbf{x}) := P(\tilde{Y} = j, K = k | \tilde{Y} \geq j, \mathbf{x}),
\label{eq:discrete_hazard_cr}
\end{equation}
the all-cause hazard as
\begin{equation}
    h(j|\mathbf{x}) := \sum_{k=1}^q h_k(j|\mathbf{x}) = P(\tilde{Y} = j | \tilde{Y} \geq j, \mathbf{x}),
\label{eq:discrete_hazard_cr_allcause}
\end{equation}
and the all-cause survival function as
\begin{equation}
    S(j|\mathbf{x}) = P(\tilde{Y} > j|\mathbf{x}) = \prod_{l = 1}^{j} (1 - h(l|\mathbf{x})).
\label{eq:discrete_survival_function_cr_allcause}
\end{equation}
From this, the CIF can be derived as 
\begin{equation}
     CIF_k(j|\mathbf{x}) = \sum_{l=1}^j P(\tilde{Y} = l, K = k|\mathbf{x})
     = \sum_{l=1}^j h_k(l|\mathbf{x}) S(l-1|\mathbf{x}).
\label{eq:discrete_cif}
\end{equation}

The data transformation necessary for competing risks analysis is as described in Section \ref{ssec:partitioning}.
For estimation of cause-specific hazards, binary classification algorithms can now simply be applied to each cause-specific dataset separately; potentially shared effects across causes can be estimated by instead stacking the $q$ datasets and using $k$ as a feature. 
The latter approach is preferred in the machine learning context, as it only requires estimation of one model and shared and cause-specific effects can be learned via interactions between feature $k$ and other features (see Section \ref{sec:applications} for an example).

A widely used DT competing risks model is the \emph{multinomial logit model} \citep{tutz2011regression, agresti2012categorical}, where the cause-specific hazard is defined as 
\begin{equation}
    h_k(j|\mathbf{x}) = \frac{\exp(\eta_{kj})}{1 + \sum_{k = 1}^q\exp(\eta_{kj})}.
\label{eq:multinomial_logit_model_hazard}
\end{equation}
This is a straightforward generalization of the \emph{logistic model} with the cause-specific linear predictor for event $k$ now defined as $\eta_{kj} = \beta_{0kj}+\mathbf{x}^{\top} \boldsymbol{\beta}_k$.
This model can further be extended to a multi-state model, by modeling all possible transitions instead of only those originating from state $0$ \citep[for details, see][]{steele2004general, tutz2016modeling}.
While the multinomial logit model has the advantage of not requiring cause-specific dataset transformations, a crucial disadvantage of this approach is that it does not lend itself to \textit{binary} classifiers anymore.

In general, just as any binary classification algorithm can be used for estimation of $h(j|\mathbf{x})$ in the single-event setting, any multiclass classification algorithm (including one vs. all reductions) can be used for estimation of $h_k(j|\mathbf{x})$ in competing risks scenarios.

\subsubsection{Limitations}
\label{sssec:limitations-dt}
DT survival methods are particularly useful in case of interval-censored survival data (e.g., caused by coarse data collection) because of their intrinsic handling of interval-censoring - provided that the intervals are identical across all subjects. 
We note that the choice of time intervals in DT models has a substantial impact on predictive accuracy, as discretization introduces a trade-off between model flexibility and information loss, making the number of bins a critical tuning hyperparameter \citep{Sloma2021EmpiricalPrediction}. However, this also depends on the choice of learner, as some methods have implicit or explicit regularization.
Individual-specific interval-censoring renders the individual-specific likelihood contributions much more complex and thus standard binary classification algorithms cannot be applied anymore \citep{tutz2016modeling}; this also holds for the case of left-censored data.
Whenever time is intrinsically continuous, however, discretization via partitioning of follow-up time causes loss of information because only time intervals as a whole are considered whereas information about the exact event time within the intervals -- as well as information about interval length -- is disregarded (as opposed to PEMs).
In addition, as for the piecewise-exponential reduction technique, the partitioning-based data transformation of the discrete-time reduction technique can substantially increase dataset size and predictions of quantities of interest other than the discrete hazard require additional post-processing.

\subsubsection{Further reading}
\label{sssec:further-reading-dt}

DT survival analysis originated as a grouped-data version of the Cox PH model with complementary log-log link \citep[\textit{Gompertz model;}][]{kalbfleischMarginalLikelihoodsBased1973}, particularly useful in case of many ties as these do not pose a problem with discretized data.
Other statistical discrete hazard models are the \emph{Probit model}, the \emph{Gumbel model}, and the \emph{exponential model} \citep[see][]{tutz2016modeling}.

Survival stacking \citep{craig2021survival} is an alternative approach to reducing survival analysis tasks to binary classification tasks, also via data transformation ("stacking"). 
Instead of explicitly discretizing follow-up time, survival stacking uses Cox-like risk sets to evaluate likelihood contributions at event time points only. 

Another popular DT reduction technique is \emph{Multi-Task Logistic Regression} \citep[\emph{MTLR}][]{yu2011learning_2}. 
The data transformation (including partitioning of follow-up time) is identical to the one presented in Section \ref{ssec:partitioning}, except that \emph{MTLR} also assigns entries to a non-censored individual $i$ after event occurrence. 
\emph{MTLR} models interval-specific event indicators jointly to enforce logical consistency across time, requiring a generalized logistic regression and limiting compatibility with standard binary classifiers. It closely resembles the DT logistic model (Eq. \ref{eq:logistic_model_hazard}) when feature effects are allowed to vary over time, effectively estimating time-varying effects.

Due to the simplicity of binary classification, discretizing survival data is hugely popular in deep learning \citep{wiegrebe.deep.2024b}. While the "natural" loss for DT survival tasks is the negative log-likelihood \citep[see][]{tutz2016modeling, zadeh2020bias, wiegrebe.deep.2024b}, the cross-entropy loss is also popular among DL-based approaches \citep[see, e.g.,][]{ren2019deep, huang2023survival} - even though it has been shwon to cause biased predictions, poor calibration, and large prediction error due to not fully exploiting the information from uncensored individuals \citep{zadeh2020bias}.

While we only considered DT approaches parametrizing the discrete hazards, others directly parametrize the probability mass function, such as \emph{TransformerJM} \citep{lin2022deep} or the popular competing risks method \emph{DeepHit} \citep{lee2018deephit}. 

\subsection{Handling of different feature and event modalities}
\label{sssec:feature-handling-pem-dt}

Due to the long format of the transformed data, time-varying features and effects can easily be incorporated into both DT and PEM survival models; for instance, \textbf{age}$_i$ (baseline age) in Table \ref{tab:data-partitioning} can simply be replaced by \textbf{age}$_{ij}$ (time-varying age) for each individual $i$ and time interval $j$.

Moreover, owing to the simplicity of the binary or Poisson response models fitted to the transformed data, more complex event modalities can easily be modeled: for instance, frailties and recurrent events (through the inclusion of random effects into the linear predictor) or competing risks and multi-state settings (through the inclusion of time-varying effects for different causes / transitions).

Importantly, $h_j(\mathbf{x})$ can represent any function of the feature vector $\mathbf{x}_{i}$, including high-order interactions. 
This enables direct modeling of complex, non-proportional hazards and time-varying effects without relying on proportionality assumptions.
Together with the above-mentioned reduction-based flexibilities, these properties make the PEM and DT approaches well-suited for machine learning algorithms.

\section{Reductions for quantity estimation at selected time points}
\label{sec:timepoint-based-reductions}

\subsection{Inverse probability of censoring weighting reduction}
\label{ssec:methods-ipcw}

The inverse probability of censoring weighting (IPCW) technique handles right-censored survival data by transforming the survival task into a weighted classification problem \citep{Vock2016}.
Specifically, IPCW allows models to predict the probability of an event (e.g., an adverse health outcome) occurring before a specific cutoff time or time horizon.
Thus, the model predicts a continuous risk score, without estimating the entire survival distribution.

\subsubsection{Single-event setting}

In the IPCW reduction, the prediction target is the conditional probability that the event occurs before or at a fixed cutoff time $\tau$, given the subject's features $\mathbf{x}_i$:

\begin{equation}
\label{eq:IPCW-pred-target}
    \pi(\mathbf{x}_i) := P(y_i \leq \tau \mid \mathbf{x}_i).
\end{equation}

The binary event label used for classification is defined as

\begin{equation}
\label{eq:IPCW-event-label}
    e_i := \mathbbm{1}(t_i \leq \tau \text{ and } d_i = 1).
\end{equation}
\newline
The IPCW procedure consists of three steps:
\begin{enumerate}
    \item Estimate the censoring survival function $\hat{G}(t)$ using the Kaplan-Meier estimator on the training data.
    \item Compute observation weights as
    \begin{equation}
        \omega_i := \frac{1}{\hat{G}(\min(t_i, \tau))},
    \end{equation}
    assigning $\omega_i = 0$ to subjects censored before $\tau$, as their status $e_i$ from Eq. \ref{eq:IPCW-event-label} is unknown.
    \item Train a binary classification algorithm using the labels $e_i$ and weights $\omega_i$ for the observations.
    This up-weights individuals with complete follow-up and down-weights those censored before $\tau$.
\end{enumerate}

The resulting model predictions $\hat{\pi}(\mathbf{x}_i)$ represent a continuous risk score: higher values indicate greater likelihood of experiencing the event before $\tau$.
These scores can also be interpreted as time-specific survival estimates via $S_i(\tau \mid \mathbf{x}_i) = 1 - \hat{\pi}(\mathbf{x}_i)$.

\subsubsection{Limitations}

IPCW can be incorporated into many standard classification models, provided that they support observation weights (e.g., classification trees, logistic regression).
Its application is less straightforward for models like neural networks, where weighting mechanisms are not inherently supported.
Moreover, the statistical validity of the method rests on the assumption that the censoring time is independent of both the event time and features.
When this assumption is violated, using unconditional Kaplan–Meier weights can potentially lead to biased predictions \citep{Gerds2006ConsistentTimes}.
Instead, more suitable approaches should estimate $\hat{G}(t|\mathbf{x})$ conditionally, using models like the Cox PH model \citep{coxRegressionModelsLifeTables1972} or flexible learners such as random survival forests \citep{ishwaranRandomSurvivalForests2008}, which can better account for feature-dependent censoring.

\subsubsection{Further reading}

Recent IPCW extensions enable handling of competing risks and dependent censoring by integrating IPCW into the resampling step rather than the learning algorithm itself, allowing the use of any standard machine learning method without modification \citep{GonzalezGinestet2021StackedRegister}.
These advances generalize previous work and improve robustness in complex survival settings.

\subsection{Complete ranking method reduction}
\label{ssec:methods-crm}

The Complete Ranking Method (CRM) addresses the challenge of handling right-censored survival data by transforming the survival task into a regression problem through a uniform ranking scheme \citep{Guan2021AAlgorithms}.
This approach is highly flexible, allowing the use of any regression algorithm, including advanced methods such as neural networks, support vector machines, and GBTs, making it broadly applicable across various domains — including survival tasks with continuous spatial or temporal data.

\subsubsection{Single-event setting}

In the CRM reduction, the prediction target is the probability that subject $i$ experiences the event of interest before a randomly chosen subject $j$:

\begin{equation}
    \label{eq:crm-pred-target}
    r(\mathbf{x}_i) := P_{j \neq i}(y_i < y_j \mid \mathbf{x}_i).
\end{equation}

CRM works by comparing all pairs of observations $(i, j)$ and computing relative risk scores $p_{ij}$ that reflect how likely it is that individual $i$ fails before individual $j$, based on their observed times $(t_i, t_j)$ and event indicators $(d_i, d_j)$.
Each pair falls into one of six possible cases, depending on the values for the event indicators and the ordering of observed times.
The score $p_{ij}$ is often computed using survival probabilities derived from the Kaplan-Meier estimator,i.e. when one of the observations is censored.

For each observation $i$, the regression target $r_i$ is calculated as the average of its pairwise scores:

\begin{equation}
\label{eq:crm-pred-label}
r_i = \frac{1}{n - 1} \sum_{j \neq i} p_{ij}.
\end{equation}

This results in a normalized target $r_i \in [0,1]$, which reflects the relative likelihood of early failure.
Higher values of $r_i$ indicate higher relative risk compared to other individuals in the dataset.

A regression model is trained to predict these targets from features, producing continuous predictions $\hat{r}(\mathbf{x}_i)$ for new observations.

\subsubsection{Limitations}

Similar to the IPCW technique utilized in \citep{Vock2016}, CRM does not directly predict the time to an event; instead, it provides a relative risk ranking of samples.
As such, its predictions focus on the likelihood of an individual failing before others rather than estimating the entire survival distribution for a subject.
Additionally, CRM has yet to be extended to handle more complex survival scenarios, such as competing risks.


\subsection{Pseudo-value reduction}
\label{ssec:methods-pv}
Pseudo-value (PV) regression is a general method to fit a regression model when a suitable nonparametric estimator of the quantity of interest is available \citep{Andersen2003}. Suppose $\widehat{\theta}$ is an unbiased estimator of a quantity of interest $\theta = \mathbb{E}(h(Y))$, where $h$ is a known function. We denote the conditional expectation by $\theta_i = \mathbb{E}(h(Y_i)| \mathbf{x}_i)$. For individual $i$, the PV is defined as    %
\begin{equation} 
\widehat{\theta}_{i} = n\widehat{\theta} - (n-1)\widehat{\theta}^{-i},
\label{eq:pv-def}
\end{equation}
where $\widehat{\theta}^{-i}$ is the value of the estimator when the $i$-th individual is removed from the dataset. PVs can be interpreted as an individual contribution to the overall estimate of the quantity of interest.  

Notably, in order to use PVs for predicting the different quantities of interest (cf. Section \ref{sec:notation}) based on individual features, we only need to replace $\hat{\theta}$ and $\hat{\theta}^{-i}$ by the respective (non-parametric) unbiased summary statistic based on the whole sample, e.g., Kaplan-Meier estimator for survival probability or Aalen-Johansen estimator for cumulative incidence function and transition probabilities. 

The main advantage of using PVs is that they can be computed for each individual at any time, regardless of censoring. Hence, by transforming survival data into PVs, methods usually restricted to uncensored data can be applied. For this reason, PVs are increasingly used in machine learning applications, as they can be treated like any standard outcome variable, enabling the direct use of learning algorithms without requiring adaptations for censored data \citep{Rahman2021, Bouaziz2023, Cwiling2023}. However, it is important to note that the PV is an asymptotically valid approximation of the quantity of interest \citep{Overgaard2017}.

Once calculated, PVs can be used as an outcome variable in a regression setting. 
In a statistical modeling context, generalized linear models are used to relate the quantity of interest to features, often estimated via generalized estimating equations (GEE; \citet{Liang1986}). 
The choice of link function and variance structure then dictates the interpretation of estimated coefficients and validity of inference. 
For an overview of PV approaches in a statistical setting, see \citet{Andersen2010}.

In most cases, PV approaches are not used to improve estimations compared to standard survival methods, but rather to provide an easier alternative in case standard survival methods involve complex modeling.
Here we consider PVs in a predictive modeling context, where machine learning models for regression can be used directly to predict the quantity of interest conditional on features (see Figures \ref{fig:tumor} and \ref{fig:icu} in Section \ref{sec:applications}).


\subsubsection{Survival probability estimation}
\label{sssec:survival-prob-estimation-pv}
Using the Kaplan-Meier estimator, we can compute PVs based on the estimated survival probability. Let $\tau$ be a specific time of interest. The PVs are defined as
\begin{equation}
\label{eq:POdef}
    {\widehat{\theta}}_i(\tau) = n{\widehat{S}(\tau)} - (n-1){\widehat{S}(\tau)}^{-i},
\end{equation}
where ${\widehat{S}(\tau)}$ is the Kaplan-Meier estimate of the survival probability at time $\tau$ and ${\widehat{S}^{-i}(\tau)}$ is the Kaplan-Meier estimator of the survival probability at time $\tau$ after removing the $i$-th individual from the dataset. 


In order to obtain survival probability estimates at different time-points ${\tau_1,\ldots, \tau_K}$, PVs are calculated at each time-point and a regression algorithm is fit to each dataset. Alternatively, the datasets can be stacked and estimated jointly. In the ML context, estimation at different time points can be achieved by stacking the PVs for each time-point and adding $\tau_k, k =1,\ldots,K$ as an additional feature.


\subsubsection{Restricted mean survival time}
\label{sssec:rmst-pv}
PVs are particularly useful in adjusting the estimation of restricted mean survival time (RMST) conditional on features. Classical survival models estimate the restricted mean survival time by first modeling the survival function and then integrating it between $0$ and $\tau$ \citep{Karrison1987, zucker1998restricted}. This approach is often computationally expensive and only gives an indirect interpretation of feature effects. On the other hand, PVs offer a direct way of regressing the RMST on features. 
For this application, one PV is defined for each subject as
\begin{equation}
\label{eq:PORMSTdef}
     {\widehat{\theta}_{i}(\tau) = n\int_0^{\tau} {\widehat{S}(t)}dt - (n-1)\int_0^{\tau} {\widehat{S}(t)}^{-i}dt}.
\end{equation}
The conditional RMST can then be estimated as 
\begin{equation}
\mathbb{E}(\min (Y_i, \tau)| \mathbf{x}_i)) = f(\mathbf{x}_i),
\end{equation}
where $f(\mathbf{x}_i)$ is a function of the features learned by the regression algorithm of choice, e.g., deep neural networks \citep{Zhao2021}, Super Learners \citep{Cwiling2024} or random forest \citep{Schenk2025}. See also Section \ref{sec:applications} for an illustration using random forests.


\subsubsection{Multi-state setting}
\label{sssec:ml-pv}

The PV approach, originally developed for modeling state probabilities in multi-state models, is broadly applicable, i.a. to the illness-death model \citep{Andersen2003} or more complex multi-state models \citep{Andersen2007}, especially where standard regression is unavailable \citep{Klein2014}.

In a multi-state setting, the quantities of interest are the transition probabilities, $P_k(\tau)$ in Equation \eqref{eq:trans-prob}. 
The Aalen-Johansen estimator \citet{ANDERSEN1996} is a suitable non-parametric estimator in this context. 
%
%
To regress transition probabilities on features, one begins by estimating the transition probability for transition $k$ at a fixed time point $\tau$ by plugging the Aalen-Johansen estimate of Equation \eqref{eq:trans-prob} into Equation \eqref{eq:pv-def}. Then, the PV for the $k$-th transition probability is given as 
\begin{equation}\label{eq:pv:MSMdef-transprob}
    \widehat{\theta}_{ki}(\tau) = n \widehat{P_k}(\tau) - (n - 1) \widehat{P_k}^{-i}(\tau),
\end{equation}
which is then again used as response variable in regression models (e.g., generalized linear models or other estimating algorithms \citep{Mogensen2013, Salerno2025}). One may want to compute multiple PVs at various times if the interest is a global estimate over time.
As the competing risks setting is a special case of the multi-state setting, Equation \eqref{eq:pv:MSMdef-transprob} still applies there, replacing the transition probabilities with an estimate of the cumulative incidence function \eqref{eq:cif}, once again using the Aalen-Johansen estimator.
This is illustrated in Section \ref{ssec:application-cif} with estimation of PV-based random forests.

\subsubsection{Limitations}

PVs are not applicable to left-truncated data \citep{Grand2019} and not easily applicable to interval-censoring. For interval-censored data, the Turnbull estimator is a non-parametric estimator of the survival probability, but PVs built from this estimator do not satisfy the asymptotic properties mentioned above \citep{Bouaziz2023}. PVs built from a parametric model for the survival function can be used \citep{Johansen2021}. The asymptotic properties of PVs hold under the assumption of fully independent censoring, which is restrictive and might not hold in practice. If censoring depends on a categorical feature, the Kaplan-Meier estimator in the PVs definition formula (\ref{eq:POdef}) can be replaced by a mixture of Kaplan-Meier estimators based on the different variable categories \citep{Andersen2010}.
One challenge associated with PVs is the arbitrary selection of time points at which PVs are defined. However, several sensitivity analyses have demonstrated that using 5 to 10 time points, equally spaced along the event time scale, typically provides sufficient information for reliable inference, see \citet{Andersen2003, Klein2005, PoharPerme2008, Andersen2010}.

\subsubsection{Further reading}
\label{sssec:further-reading-pv}
Recent developments in multi-state models using PVs include the extension to interval-censored data \citep{Sabathe2020} and the relaxation of the Markov assumption \citep{Andersen2022}. PVs are also used in relative survival, where they offer an alternative that is, easier to implement in software than the existing estimator \citep{Pavlic2019}. PVs are also useful to implement cure models \citep{Su2022} and can be used as graphical tools to check model assumptions for hazard regression models (Cox model, additive model, Fine and Gray model), see \citet{PoharPerme2008}. 

\section{Software}
\label{sec:software}

Several packages provide standalone implementations of specific reduction techniques for survival analysis. Notable examples include \texttt{pammtools} \citep{bender2018pammtools}, which provides functions for the necessary data transformations for piecewise exponential reductions, \texttt{discSurv} \citep{diskSurv_Rpackage} for discrete-time data transformation, and \texttt{pseudo} \citep{pseudoRpackage} for calculation of pseudo values in different settings.
These implementations typically focus on data transformation or specialized workflows in a statistical modeling context rather than machine learning workflows.

The reduction framework described in Figure~\ref{fig:reduction-pipeline} aligns naturally with the design philosophy of the \texttt{mlr3pipelines} package \citep{mlr3pipelines2021}, which modularizes machine learning workflows into reusable and composable building blocks (\texttt{PipeOps}), each with well-defined train and predict semantics, standardized input/output interfaces, and internal state management.
Leveraging this architecture, we implemented several reduction techniques within the \texttt{mlr3proba} R package \citep{Sonabend2021} for right-censored survival data.
Specifically, we provide implementations for the PEM (Section~\ref{ssec:methods-pem}), DT (Section~\ref{ssec:methods-dt}), and IPCW reduction (Section~\ref{ssec:methods-ipcw}).

Each method is implemented as a self-contained pipeline consisting of two main components: a train PipeOp that performs the appropriate data transformation during training (e.g., long-format conversion for DT or PEM) and stores relevant parameters in its internal state (e.g., the cut-points or time grid), and a prediction PipeOp that maps predictions from a regression or classification model back to the survival domain using the internal state.

This design ensures full compatibility with the broader \textit{mlr3} ecosystem, enabling flexible model selection across regression, classification, and survival learners, integrated resampling and hyperparameter tuning with nested cross-validation, and support for internal validation and early stopping \citep{Fischer2024_validation}, such as with XGBoost and PEM.
Furthermore, transformed tasks retain the original subject identifiers, ensuring that resampling and evaluation operate on the correct observational units, rather than individual rows in the long-format data. This procedure avoids bias from pseudo-replication due to multiple rows per subject in the long-format data.
This makes reductions compatible with standard performance metrics and resampling schemes in survival analysis.
The pipeline design also allows high flexibility.
For instance, with the PEM pipeline, users can explicitly specify the time-varying design formula (e.g., including interval end-times as features to capture time-varying effects) and can also directly control the discretization of the follow-up time.

By building on the modularity of the \textit{mlr3} framework \citep{Lang2019}, our implementations offer a unified, extensible, and reproducible interface to reduction-based survival modeling.
They enable rapid prototyping, tuning, and deployment of survival models using familiar tools from the ML ecosystem—closing the gap between modern ML methods and classical SA requirements.

While the reduction pipelines implemented in \texttt{mlr3proba} are technically applicable to left-truncated data, competing risks and multi-state settings (i.e., data transformation and model estimation work), automated evaluation is currently not supported as this requires a refactoring of the container that stores model predictions. However, this extension will be available in future releases.



\section{Applications}
\label{sec:applications}
In this section, we showcase applications of the PEM, DT, and PV reduction techniques in a single-event and a competing risks setting, using XGBoost for PEM \citep{chen2016xgboost} and random forests for DT and PV \citep{breiman2001random} as estimation algorithms.
In doing so, we aim to demonstrate similarities and differences across these reduction techniques as well as their flexibility in accommodating distinct survival tasks and machine learning algorithms. While the chosen examples are deliberately simple (low-dimensional, no hyperparameter tuning, no out of sample evaluation), they illustrate that the respective approaches are in principle able to learn the underlying event time distribution, even in the presence of non-proportional hazards.
A quantitative evaluation of the DT and PEM-based reductions is given in Section \ref{sec:benchmarks}.

\subsection{Estimation of the survival function and RMST in a single-event setting}
\label{ssec:application-surv-rmst}

In the single-event setting, we illustrate the estimation of survival function and RMST on the \texttt{tumor} dataset, included in the \textsf{R} package \texttt{pammtools} \citep{bender2018pammtools}. The \texttt{tumor} dataset contains information on 776 patients treated for a tumor located in the stomach area. The outcome of interest is time from tumor surgery until death ($1-3,000$ days). In our illustrations, we consider the binary variable \texttt{complications}, which indicates whether or not major complications occurred during surgery.

For the PV reduction technique, the PVs were independently calculated at each time of interest and based on the non-parametric KM estimator following Equation \eqref{eq:POdef}.
For the single-event \texttt{tumor} dataset, the data transformation required for partitioning-based reduction techniques (PEM and DT) is precisely as illustrated in Section \ref{ssec:partitioning} (in particular, Figure \ref{fig:fu-partitioning} and Table \ref{tab:data-partitioning}).

\begin{figure}[!ht]
    \centering
    \includegraphics[width=1\linewidth]{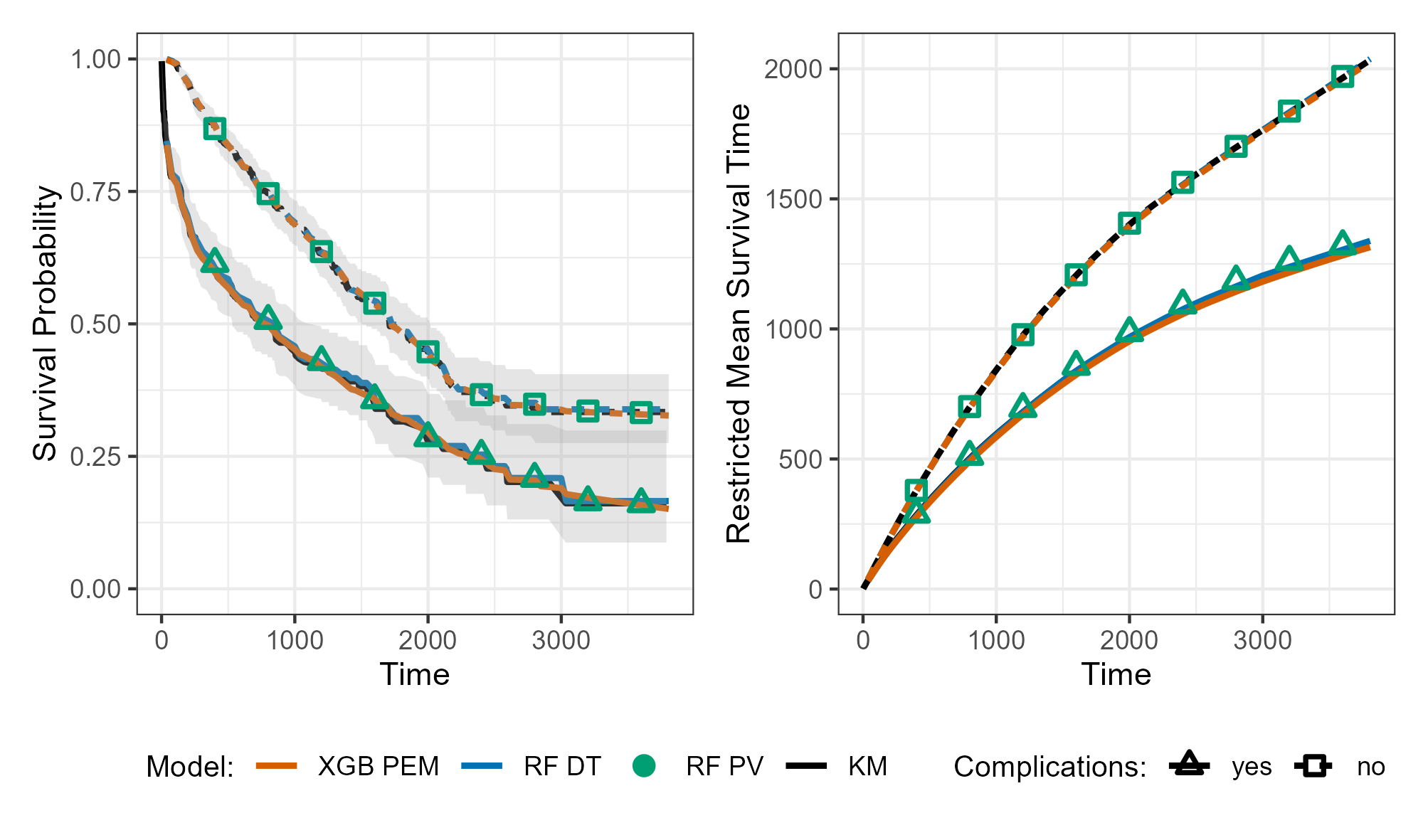}
    \vspace{-0.4cm}
    \caption{Comparison of survival probabilities and RMST across reduction techniques for the \texttt{tumor} dataset. 
    This figure shows predicted survival probabilities (upper panel) and RMST (lower panel) stratified by complication status, for the three reduction techniques PEM, DT, and PV, as well as for the non-parametric Kaplan-Meier (KM) estimator. 
    For PEM, XGBoost was employed; for DT and PV, random forests were applied.
    For the DT and PEM approaches and the Kaplan-Meier estimator, curves of the survival function and RMST are shown; for the PV approach, point estimates at specific time points are displayed for both survival probabilities and RMST.}
    \label{fig:tumor}
\end{figure}

Figure \ref{fig:tumor} illustrates predicted survival probabilities (left panel) and the RMST (right panel) across time and stratified by complication status. 
Prediction curves are displayed for the DT and PEM reduction techniques (and for the Kaplan-Meier estimator as "ground truth"), whereas for the PV approach, only point estimates are shown, underlining the fact that the PV reduction is usually used for evaluating quantities of interest at specific time points.
For the PV reduction technique, we use regression random forest for the computation of predicted survival probabilities and RMST, based on Equations \eqref{eq:POdef} \eqref{eq:PORMSTdef}, respectively. 
We use XGBoost \citep{chen2016xgboost} for the PEM-based estimates (as they implement Poisson likelihood and allow inclusion of an offset) and random forest for classification for the DT approach. 

Notably, all approaches recover the Kaplan-Meier estimates very well, although the survival curves for the two groups (complications yes vs. no) are clearly non-proportional. 
This means that XGBoost and RF successfully learned the interaction of the feature time and complications. 

While predicting survival probability curves is straightforward with the DT and PEM methods, calculating the RMST is, in general, not. 
Here we used the method by \citep{zucker1998restricted}, which becomes very complicated as more (especially continuous) features are added to the model. On the contrary, PVs are particularly useful in adjusting the estimation of restricted mean survival time conditional on features because they enable the modeling of the relation between the mean survival time and given features by a direct regression model\citep{Andersen2004}. 

\subsection{Estimation of cumulative incidence}
\label{ssec:application-cif}

In the competing risks setting, we show the estimation of CIFs using the \texttt{sir.adm} dataset \citep{beyersmann2006use}. The dataset comprises a random subsample of 747 patients from the prospective SIR 3 (Spread of nosocomial Infections and Resistant
pathogens) cohort study at the Charit\'e university hospital in Berlin, Germany, which aimed to investigate the effect of hospital-acquired infections in intensive care \citep{wolkewitz2008risk}. 
It contains information on patients' pneumonia status at admission to the intensive care unit (ICU), time of ICU stay, and whether the patient was discharged alive or died in the hospital.

The necessary data transformation for partitioning-based reduction techniques (PEM and DT) is described in Section \ref{ssec:partitioning} and illustrated for an excerpt of the \texttt{sir.adm} dataset in Table \ref{tab:data-partitioning-dt-cr}. 

\begin{table}[!ht]
\hfill
\begin{tabular}{|c|c|c|c|c|c|c|c|c|c|c|}
\hline
\textbf{i} & \textbf{j} & \textbf{I}$_j$ & $\textbf{d}_{ijk}$ & $\textbf{t}_{ij}$ & $\textbf{o}_{ij}$ & $\textbf{a}_{j}$ & $\textbf{pneu}_i$ & $\textbf{age}_i$ & $\textbf{sex}_i$ & \textbf{cause} \\ \hline
30238 & 1   & (0,2]   & 0 & 2 & 0.7 & 2  & 1 & 38.7 & M & 1 \\ \hline
30238 & 25  & (25,26] & 0 & 1 & 0   & 26 & 1 & 38.7 & M & 1 \\ \hline
41    & 1   & (0,2]   & 0 & 2 & 0.7 & 2  & 0 & 75.3 & F & 1 \\ \hline
41    & 3   & (3,4]   & 1 & 1 & 0   & 4  & 0 & 75.3 & F & 1 \\ \hline
17058 & 1   & (0,2]   & 0 & 2 & 0.7 & 2  & 1 & 62.2 & M & 1 \\ \hline
17058 & 21  & (21,22] & 0 & 1 & 0   & 22 & 1 & 62.2 & M & 1 \\ \hline
\end{tabular} \\
\vspace{0.1cm}

\hspace{1cm}
$k=1$ (discharge) \scalebox{2}{$\nearrow$} \\[1cm]

\vspace{-0.9cm}
\begin{tabular}{|c|c|c|c|c|c|}
\hline
\textbf{i} & \textbf{k} & $\textbf{t}_i$ & $\textbf{pneu}_i$ & $\textbf{age}_i$ & $\textbf{sex}_i$ \\ \hline
30238 & 0 & & 26 1 & 38.7 & M \\ \hline
41 & 1 & 75.3 & 0 & 4 & F \\ \hline
17058 & 2 & 62.2 & 1 & 22 & M \\ \hline
\end{tabular}
\vspace{0.1cm}

\hspace{1cm}
$k=2$ (death) \hspace{0.4cm} \scalebox{2}{$\searrow$}

\vspace{0.1cm}

\hfill
\begin{tabular}{|c|c|c|c|c|c|c|c|c|c|c|}
\hline
\textbf{i} & \textbf{j} & \textbf{I}$_j$ & $\textbf{d}_{ijk}$ & $\textbf{t}_{ij}$ & $\textbf{o}_{ij}$ & $\textbf{a}_{j}$ & $\textbf{pneu}_i$ & $\textbf{age}_i$ & $\textbf{sex}_i$ & \textbf{cause} \\ \hline
30238 & 1   & (0,2]   & 0 & 2 & 0.7 & 2  & 1 & 38.7 & M & 2 \\ \hline
30238 & 25  & (25,26] & 0 & 1 & 0   & 26 & 1 & 38.7 & M & 2 \\ \hline
41    & 1   & (0,2]   & 0 & 2 & 0.7 & 2  & 0 & 75.3 & F & 2 \\ \hline
41    & 3   & (3,4]   & 0 & 1 & 0   & 4  & 0 & 75.3 & F & 2 \\ \hline
17058 & 1   & (0,2]   & 0 & 2 & 0.7 & 2  & 1 & 62.2 & M & 2 \\ \hline
17058 & 21  & (21,22] & 1 & 1 & 0   & 22 & 1 & 62.2 & M & 2 \\ \hline
\end{tabular}
\caption{Illustration of data transformation for competing risks analysis with the \texttt{sir.adm} dataset. 
Starting from the \texttt{sir.adm} dataset in standard format --- i.e., with a single row per individual $i$ containing the ID, cause (0: censoring; 1: discharge; 2: death), observed time, and features - the follow-up time is partitioned into time intervals as illustrated in Figure \ref{fig:fu-partitioning}. 
Subsequently, expanded long-format datasets with one row per individual and time interval are created separately for each cause $k \in \{1,2\}$.
This works analogously to Table \ref{tab:data-partitioning}, the only difference being that the event indicator is now $d_{ijk}$, indicating whether individual $i$ experiences an event of type $k$ in interval $j$. 
In addition, for each cause-specific dataset, a column denoting the cause is added.} 
\label{tab:data-partitioning-dt-cr}
\end{table}

Importantly, the resulting cause-specific datasets are stacked for model estimation of partition-based reduction techniques (cf. Section \ref{sssec:competing-risk-dt}). 
For the PV approach, survival data are transformed into PVs for each cause using the non-parametric Aalen-Johansen estimator, following equation \eqref{eq:pv:MSMdef-transprob}.

\begin{figure}[!ht]
    \centering
    \includegraphics[width=1\linewidth]{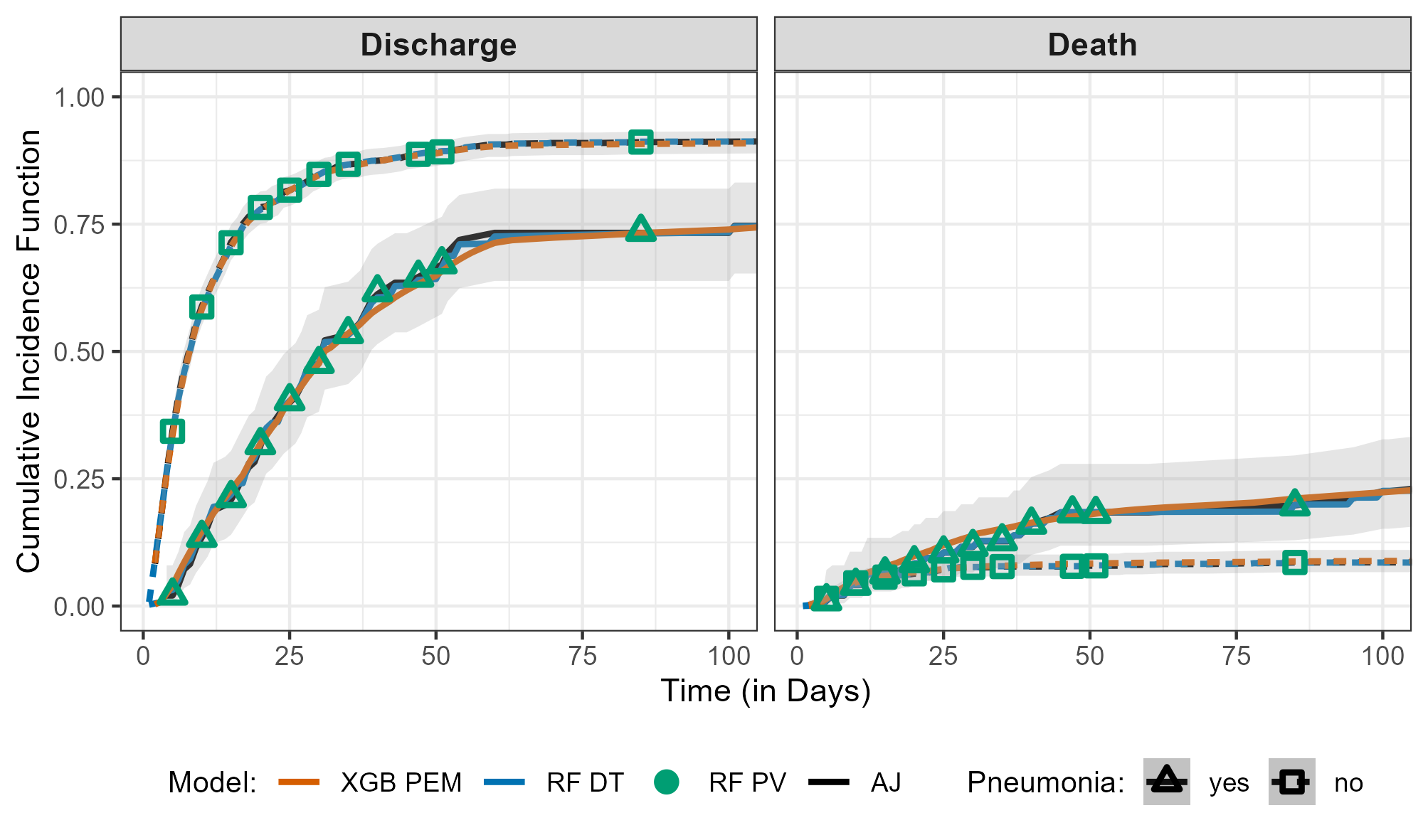}
    \vspace{-0.4cm}
    \caption{Comparison of cumulative incidence across reduction techniques for the competing risks \texttt{sir.adm} dataset. 
    This figure shows predicted survival probabilities by event type and stratified by complication status, for the three reduction techniques DT, PEM, and PV as well as for the non-parametric Aalen-Johansen (AJ) estimator. 
    For PEM, XGBoost (XGB PEM) was employed; for DT and PV, random forests (RFC DT/PV) were applied.
    For the DT and PEM approaches and the Aalen-Johansen estimator, curves of the survival function are shown, whereas for the PV approach, point estimates of the survival probability at specific time points are shown.}
    \label{fig:icu}
\end{figure}

Figure \ref{fig:icu} illustrates predicted cumulative incidence (Eq. \eqref{eq:cif}) probabilities by cause, across time, and stratified by pneumonia status. 
As in the single-event setting, predicted curves are displayed for the DT and PEM reduction techniques (and the non-parametric baseline Aalen-Johansen estimator); for PV, point estimates at specific time points are shown.
The PVs are used as the outcome variable to estimate the cumulative incidences from a random forest regression tree.

\section{Benchmarking experiments}
\label{sec:benchmarks}

We conduct a benchmark experiment to compare the predictive performance of multiple reduction techniques with common survival learners using the \textit{mlr3} framework and our reduction implementation detailed in Section \ref{sec:software}. We focus on reduction techniques that provide an estimate of the event time distribution, namely PEM and DT reductions. 
The experiment comprises nine tasks with seven real-world datasets available in R packages from CRAN and two synthetic datasets simulated to illustrate specific challenges in predictive modeling for survival tasks (see Table \ref{tab:bm-tasks} in Appendix \ref{sec:appendix-a}). \texttt{synthetic-breakpoint} simulated data with a change point in the baseline hazard. \texttt{synthetic-tve} contains simulated data with highly non-linear time-varying effects (i.e. non-proportional hazards). For the general setup of the benchmark, including choice of hyperparameter space, we followed the large scale benchmark in \citet{burk2024largescaleneutral}.
The goal of the benchmark in this paper was not to perform an exhaustive large-scale comparison of different learners, but rather to illustrate that predictive performance of reduction techniques is largely on par with survival-specific learners.

The learners used for comparison include Kaplan-Meier (KM) as a baseline, $L_2$-penalized Cox regression (RIDGE), the Cox elastic net (GLMN), random survival forest (RSFRC) and classification random forest with discrete-time reduction (RSFRC\_DT) as well as XGBoost with Cox objective (XGBCox), piecewise exponential reduction (XGB\_PEM), and discrete-time reduction (XGB\_DT) (see Table \ref{tab:bm-learners} in Appendix \ref{sec:appendix-a}).

We use a nested resampling tuning \citep{bischl2023hpo} procedure using 3-fold cross-validation for inner resampling and repeated 3-fold cross-validation with a variable number of repetitions, using three repetitions for tasks with fewer events ($\leq 500$ events), one iteration for tasks with over 1000 events, and 2 repetitions otherwise.
Hyperparameters are tuned using Bayesian optimization \citep{garnett2023bayesoptbook} using a hybrid tuning budget of either a set number of evaluations scaling with the number of tunable parameters per learner $50 \cdot n_\theta$ or until a time limit of 120 hours (five days) was reached.
All variants of XGBoost use early stopping for the \texttt{nrounds} parameter (stopping after 50 iterations without improvement).

The experiment is conducted twice in total, once using Harrell's C-index for tuning and evaluation \citep{Harrell1982}, and once using the integrated survival brier score \citep[ISBS;][]{Graf1999}) analogously, where we integrate up to the 80th percentile of observed times in each training set \citep{Sonabend2022}.
To ensure consistency and fairness of our results, we employ a KM as a fallback learner \citep[e.g.][]{mlr3booklargescale} which is used to impute performance scores in cases where the learner in question can not produce a result due to underlying errors, such as numerical issues or other implementation-specific reasons.
This also results in learners with many errors producing results skewed towards the performance of the baseline KM results, making stability an indirect additional evaluation metric.

Results are presented aggregated per learner across all tasks in Figure \ref{fig:bm-aggr} and in Appendix \ref{sec:appendix-a} (Table \ref{tab:bm-aggr}), as well as separately for each task aggregated across outer resampling iterations in Appendix \ref{sec:appendix-a} (Table \ref{tab:bm-scores}, Figures \ref{fig:bm-scores-harrell-c} and \ref{fig:bm-scores-isbs}), where we also include tables of evaluation scores.
The results indicate that overall the reduction based approaches have competitive performance to survival-specific learners. However, the performance appears to depend to some extent on the underlying classification/regression learner. For example, while XGBoost based DT reduction performs quite well, while the random forest based DT reduction underperforms on many tasks compared to random survival forest. On the other hand, both PEM and DT reductions with XGBoost outperform the XGBoost based Cox implementation, which frequently errored during estimation and thus appears to be less stable compared to the respective reductions (in its current implementation).

\begin{figure}
     \centering
     \begin{subfigure}[b]{0.9\textwidth}
         \centering
         \includegraphics[width=\textwidth]{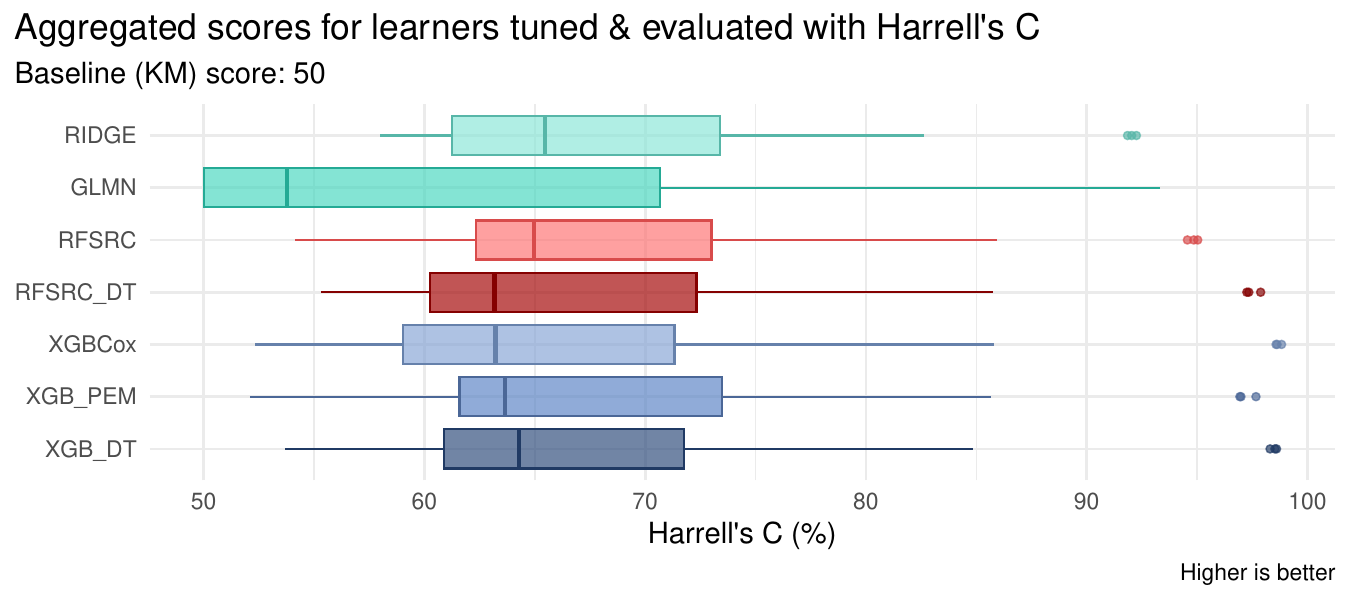}
         \caption{Harrell's C-index scores aggregated across all tasks, scaled by 100.}
         \label{fig:bm-aggr-harrell-c}
     \end{subfigure}
     
     \begin{subfigure}[b]{0.9\textwidth}
         \centering
         \includegraphics[width=\textwidth]{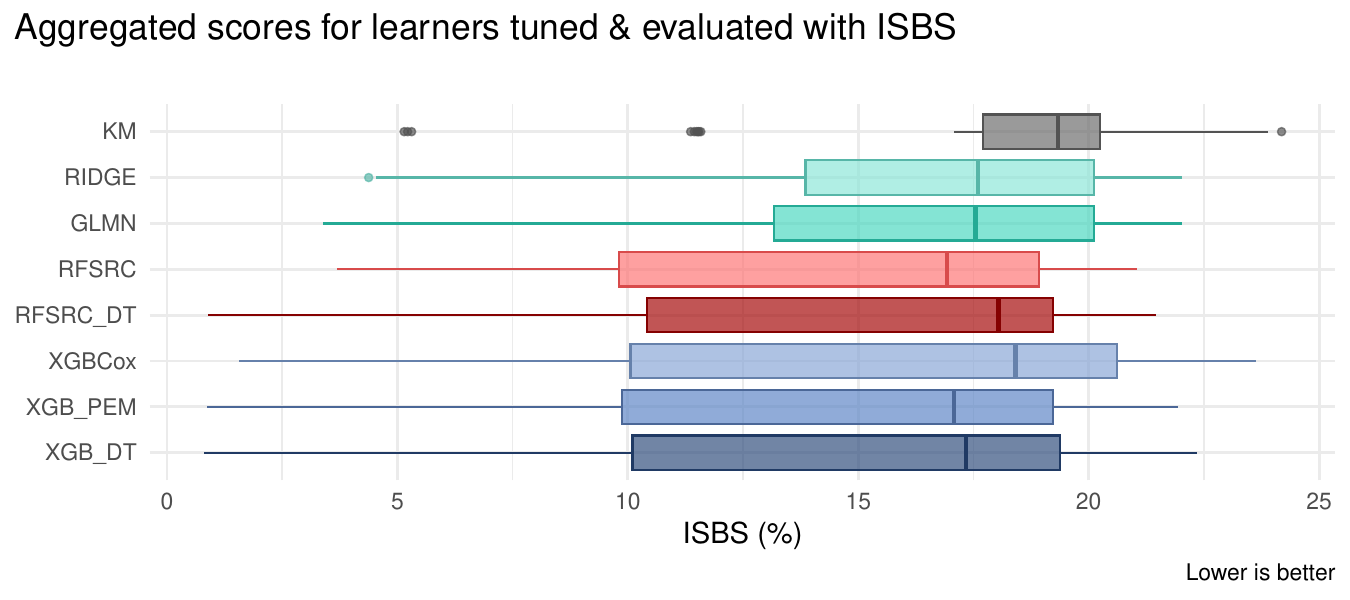}
         \caption{Integrated Survival Brier Score (ISBS) aggregated across all tasks, scaled by 100}
         \label{fig:bm-aggr-isbs}
     \end{subfigure}
        \caption{Aggregated benchmark results across all tasks presented as boxplots.}
        \label{fig:bm-aggr}
\end{figure}

\section{Discussion}
\label{sec:discussion}

In this work, we proposed a general strategy for reducing complex survival analysis tasks to standard regression or classification tasks -- which in turn enables the application of standard machine and deep learning methods -- and provided an overview of such reduction techniques.
Overall, these reduction techniques account for the particular incompleteness of information in survival data (e.g., right-censoring and left-truncation), are applicable to a broad range of survival tasks (e.g., competing risks or multi-state settings), can predict different quantities of interest (e.g., hazards, CIFs), and can use off-the-shelf implementations of machine or deep learning methods (e.g., boosting, random forest), while not imposing (strong) assumptions regarding the underlying distribution of event times.
Our implementation in \texttt{mlr3proba} provides a principled way to make such reductions available for practitioners. Beyond the specific implementation, we provide a blueprint for the implementaiton of such techniques in general ML workflows.

Empirical results from selected data examples and a benchmark study on multiple real-world and synthetic datasets indicate that standard machine learning algorithms for regression and classification in combination with reduction techniques are competitive with established survival-specific learners.  

Despite their general applicability, each reduction technique also comes with its own limitations. 
For instance, PEMs and DT models are not applicable to left-censored data, while PV-based approaches are not applicable to left-truncated data. None of the techniques can easily incorporate individual-specific interval-censored data.
At the same time, while our current benchmark experiments were comprehensive, they could be more exhaustive regarding hyperparameter tuning and choice of learners.
Currently, the implementation of reduction techniques within \texttt{mlr3proba} is limited to PEM, DT and IPCW approaches, with automated evaluation not yet supported for many survival tasks beyond single-event, right-censoring data.

Future work will look at other reduction techniques as well as at extensions of the existing ones to survival settings currently not covered. In particular, PV based reductions will be integrated in \texttt{mlr3proba}.
Moreover, further empirical evaluation of different base learner/reduction technique combinations could elicit a better understanding of the underlying processes and reasons for differences in performance.

\newpage

\section*{Data availability}
All datasets used in this work (Sections \ref{sec:applications} and \ref{sec:benchmarks}) are public datasets, accessible via the GitHub repositories \href{https://github.com/survival-org/reduction-techniques}{https://github.com/survival-org/reduction-techniques} and \href{https://github.com/survival-org/reduction-techniques-benchmark}{https://github.com/survival-org/reduction-techniques-benchmark}.

\section*{Code availability}
All code required for running the application and benchmark analyses in this work (Sections \ref{sec:applications} and \ref{sec:benchmarks}), as well as for creating the corresponding figures, is available from the GitHub repositories \href{https://github.com/survival-org/reduction-techniques}{https://github.com/survival-org/reduction-techniques} and \href{https://github.com/survival-org/reduction-techniques-benchmark}{https://github.com/survival-org/reduction-techniques-benchmark}.


\clearpage

\appendix
\section{Appendix}\label{sec:appendix-a}

\subsection{Additional benchmark details and results}
\FloatBarrier
\begin{table}[h]

\caption{Tasks used in benchmark comparison including number of observations (N), features (p), and observed events, the censoring rate, and number of repeats for outer repeated cross-validation used for evaluation.\label{tab:bm-tasks}}
\centering
\begin{tabular}[t]{lrrrrr}
\toprule
Task & N & p & Events & Cens. \% & Repeats\\
\midrule
cat\_adoption & 2257 & 18 & 1434 & 36.46 & 1\\
mgus & 176 & 7 & 165 & 6.25 & 3\\
nafld1 & 12588 & 5 & 1018 & 91.91 & 1\\
nwtco & 4028 & 3 & 571 & 85.82 & 2\\
std & 877 & 21 & 347 & 60.43 & 3\\
synthetic-breakpoint & 2000 & 20 & 1108 & 44.60 & 1\\
synthetic-tve & 2000 & 24 & 1508 & 24.60 & 1\\
tumor & 776 & 7 & 375 & 51.68 & 3\\
wa\_churn & 7032 & 18 & 1869 & 73.42 & 1\\
\bottomrule
\end{tabular}
\end{table}

\begin{table}[h]

\caption{Learners used in the benchmark comparison, including their \textit{mlr3} IDs, implementing R packages, and the number of explicitly tuned hyperparameters. Note \texttt{cv.glmnet} internally tunes the \texttt{lambda} regularization parameter. \label{tab:bm-learners}}
\centering
\begin{tabular}[h]{lllr}
\toprule
ID & Package & mlr3 ID & \# Parameters\\
\midrule
KM & survival & surv.kaplan & 0\\
RIDGE & glmnet & surv.cv\_glmnet & 0\\
GLMN & glmnet & surv.cv\_glmnet & 1\\
RFSRC & randomForestSRC & surv.rfsrc & 5\\
RFSRC\_DT & randomForestSRC & classif.rfsrc & 5\\
XGBCox & xgboost & surv.xgboost.cox & 5\\
XGB\_PEM & xgboost & regr.xgboost & 5\\
XGB\_DT & xgboost & classif.xgboost & 5\\
\bottomrule
\end{tabular}
\end{table}

\begin{table}[h]
\centering
\caption{Mean (SD) of evaluation scores aggregated by learner. Scores scaled by 100 for readability.\label{tab:bm-aggr}}
\centering
\begin{tabular}[t]{lll}
\toprule
Learner & Harrell's C & ISBS\\
\midrule
KM & 50 (0) & 17.87 (4.56)\\
RIDGE & 68.67 (9.51) & 16.04 (5.3)\\
GLMN & 61.89 (14.12) & 15.86 (5.45)\\
RFSRC & 69.11 (10.28) & 14.78 (5.52)\\
RFSRC\_DT & 67.98 (11.36) & 15.23 (6.04)\\
XGBCox & 67.12 (12.31) & 15.81 (6.51)\\
XGB\_PEM & 68.4 (11.31) & 14.79 (6.04)\\
XGB\_DT & 68.19 (11.44) & 14.99 (6.18)\\
\bottomrule
\end{tabular}
\end{table}

\clearpage

\begin{longtable}[t]{lll}
\caption{Mean (SD) of evaluation scores for each learner and task. Scores scaled by 100 for readability.\label{tab:bm-scores}}\\
\toprule
Learner & Harrell's C & ISBS\\
\midrule
\endfirsthead
\caption[]{Mean (SD) of evaluation scores for each learner and task. Scores scaled by 100 for readability. \textit{(continued)}}\\
\toprule
Learner & Harrell's C & ISBS\\
\midrule
\endhead

\endfoot
\bottomrule
\endlastfoot
\addlinespace[0.3em]
\multicolumn{3}{l}{\textbf{synthetic-breakpoint}}\\
\hspace{1em}\cellcolor{gray!10}{KM} & \cellcolor{gray!10}{50 (0)} & \cellcolor{gray!10}{18.09 (0.36)}\\
\hspace{1em}RIDGE & 59.5 (1.27) & 18.05 (0.3)\\
\hspace{1em}\cellcolor{gray!10}{GLMN} & \cellcolor{gray!10}{53.55 (6.15)} & \cellcolor{gray!10}{18.09 (0.36)}\\
\hspace{1em}RFSRC & 63 (0.88) & 16.64 (0.25)\\
\hspace{1em}\cellcolor{gray!10}{RFSRC\_DT} & \cellcolor{gray!10}{62 (1.45)} & \cellcolor{gray!10}{18.09 (0.73)}\\
\hspace{1em}XGBCox & 61.53 (2.17) & 16.77 (0.48)\\
\hspace{1em}\cellcolor{gray!10}{XGB\_PEM} & \cellcolor{gray!10}{62.68 (0.29)} & \cellcolor{gray!10}{16.53 (0.39)}\\
\hspace{1em}XGB\_DT & 63.1 (1.37) & 16.8 (0.45)\\
\addlinespace[0.3em]
\multicolumn{3}{l}{\textbf{synthetic-tve}}\\
\hspace{1em}\cellcolor{gray!10}{KM} & \cellcolor{gray!10}{50 (0)} & \cellcolor{gray!10}{23.91 (0.26)}\\
\hspace{1em}RIDGE & 80.02 (0.83) & 16.92 (0.38)\\
\hspace{1em}\cellcolor{gray!10}{GLMN} & \cellcolor{gray!10}{80.7 (0.79)} & \cellcolor{gray!10}{14.39 (0.85)}\\
\hspace{1em}RFSRC & 84.47 (1.53) & 9.25 (0.31)\\
\hspace{1em}\cellcolor{gray!10}{RFSRC\_DT} & \cellcolor{gray!10}{85.25 (0.67)} & \cellcolor{gray!10}{11.82 (1.17)}\\
\hspace{1em}XGBCox & 85.46 (0.41) & 10.15 (0.47)\\
\hspace{1em}\cellcolor{gray!10}{XGB\_PEM} & \cellcolor{gray!10}{84.79 (0.77)} & \cellcolor{gray!10}{9.55 (0.41)}\\
\hspace{1em}XGB\_DT & 84.28 (0.5) & 9.5 (0.46)\\
\addlinespace[0.3em]
\multicolumn{3}{l}{\textbf{cat\_adoption}}\\
\hspace{1em}\cellcolor{gray!10}{KM} & \cellcolor{gray!10}{50 (0)} & \cellcolor{gray!10}{21.46 (0.6)}\\
\hspace{1em}RIDGE & 59.98 (0.43) & 21.46 (0.6)\\
\hspace{1em}\cellcolor{gray!10}{GLMN} & \cellcolor{gray!10}{52.52 (4.37)} & \cellcolor{gray!10}{21.46 (0.6)}\\
\hspace{1em}RFSRC & 62.05 (0.48) & 20.36 (0.79)\\
\hspace{1em}\cellcolor{gray!10}{RFSRC\_DT} & \cellcolor{gray!10}{59.6 (2.36)} & \cellcolor{gray!10}{21.01 (0.74)}\\
\hspace{1em}XGBCox & 61.94 (1.02) & 21.59 (1.15)\\
\hspace{1em}\cellcolor{gray!10}{XGB\_PEM} & \cellcolor{gray!10}{61.21 (0.74)} & \cellcolor{gray!10}{20.93 (1.23)}\\
\hspace{1em}XGB\_DT & 60.89 (0.37) & 21.02 (1.1)\\
\addlinespace[0.3em]
\multicolumn{3}{l}{\textbf{wa\_churn}}\\
\hspace{1em}\cellcolor{gray!10}{KM} & \cellcolor{gray!10}{50 (0)} & \cellcolor{gray!10}{17.32 (0.32)}\\
\hspace{1em}RIDGE & 92.05 (0.2) & 4.56 (0.16)\\
\hspace{1em}\cellcolor{gray!10}{GLMN} & \cellcolor{gray!10}{93.12 (0.2)} & \cellcolor{gray!10}{3.56 (0.17)}\\
\hspace{1em}RFSRC & 94.81 (0.24) & 3.84 (0.14)\\
\hspace{1em}\cellcolor{gray!10}{RFSRC\_DT} & \cellcolor{gray!10}{97.49 (0.34)} & \cellcolor{gray!10}{1.01 (0.14)}\\
\hspace{1em}XGBCox & 98.68 (0.13) & 1.7 (0.14)\\
\hspace{1em}\cellcolor{gray!10}{XGB\_PEM} & \cellcolor{gray!10}{97.2 (0.41)} & \cellcolor{gray!10}{0.92 (0.09)}\\
\hspace{1em}XGB\_DT & 98.48 (0.15) & 0.86 (0.08)\\
\addlinespace[0.3em]
\multicolumn{3}{l}{\textbf{std}}\\
\hspace{1em}\cellcolor{gray!10}{KM} & \cellcolor{gray!10}{50 (0)} & \cellcolor{gray!10}{19.92 (0.71)}\\
\hspace{1em}RIDGE & 60.13 (1.73) & 19.92 (0.71)\\
\hspace{1em}\cellcolor{gray!10}{GLMN} & \cellcolor{gray!10}{50 (0)} & \cellcolor{gray!10}{19.92 (0.71)}\\
\hspace{1em}RFSRC & 59.36 (2.79) & 19.29 (0.99)\\
\hspace{1em}\cellcolor{gray!10}{RFSRC\_DT} & \cellcolor{gray!10}{58.19 (1.84)} & \cellcolor{gray!10}{19.69 (0.78)}\\
\hspace{1em}XGBCox & 55.74 (2.63) & 21.29 (1.86)\\
\hspace{1em}\cellcolor{gray!10}{XGB\_PEM} & \cellcolor{gray!10}{57.11 (2.9)} & \cellcolor{gray!10}{19.71 (1.09)}\\
\hspace{1em}XGB\_DT & 57.58 (1.9) & 20.37 (1.04)\\
\addlinespace[0.3em]
\multicolumn{3}{l}{\textbf{mgus}}\\
\hspace{1em}\cellcolor{gray!10}{KM} & \cellcolor{gray!10}{50 (0)} & \cellcolor{gray!10}{19.01 (0.52)}\\
\hspace{1em}RIDGE & 70.18 (4.65) & 16.15 (0.9)\\
\hspace{1em}\cellcolor{gray!10}{GLMN} & \cellcolor{gray!10}{69.65 (3.58)} & \cellcolor{gray!10}{16.34 (0.81)}\\
\hspace{1em}RFSRC & 68.71 (4.14) & 15.87 (1.41)\\
\hspace{1em}\cellcolor{gray!10}{RFSRC\_DT} & \cellcolor{gray!10}{66.3 (4)} & \cellcolor{gray!10}{16.65 (1.77)}\\
\hspace{1em}XGBCox & 64.49 (5.22) & 17.88 (1.72)\\
\hspace{1em}\cellcolor{gray!10}{XGB\_PEM} & \cellcolor{gray!10}{68.3 (4.71)} & \cellcolor{gray!10}{16.11 (1.57)}\\
\hspace{1em}XGB\_DT & 67.83 (3.59) & 16.36 (1.69)\\
\addlinespace[0.3em]
\multicolumn{3}{l}{\textbf{nafld1}}\\
\hspace{1em}\cellcolor{gray!10}{KM} & \cellcolor{gray!10}{50 (0)} & \cellcolor{gray!10}{5.23 (0.08)}\\
\hspace{1em}RIDGE & 82.52 (0.11) & 4.6 (0.1)\\
\hspace{1em}\cellcolor{gray!10}{GLMN} & \cellcolor{gray!10}{82.58 (0.08)} & \cellcolor{gray!10}{4.59 (0.15)}\\
\hspace{1em}RFSRC & 83.22 (0.45) & 4.11 (0.03)\\
\hspace{1em}\cellcolor{gray!10}{RFSRC\_DT} & \cellcolor{gray!10}{82.96 (0.15)} & \cellcolor{gray!10}{4.21 (0.06)}\\
\hspace{1em}XGBCox & 83.11 (0.05) & 4.13 (0.05)\\
\hspace{1em}\cellcolor{gray!10}{XGB\_PEM} & \cellcolor{gray!10}{82.95 (0.14)} & \cellcolor{gray!10}{4.14 (0.04)}\\
\hspace{1em}XGB\_DT & 82.96 (0.12) & 4.11 (0.03)\\
\addlinespace[0.3em]
\multicolumn{3}{l}{\textbf{nwtco}}\\
\hspace{1em}\cellcolor{gray!10}{KM} & \cellcolor{gray!10}{50 (0)} & \cellcolor{gray!10}{11.49 (0.08)}\\
\hspace{1em}RIDGE & 70.99 (2.28) & 11.33 (0.11)\\
\hspace{1em}\cellcolor{gray!10}{GLMN} & \cellcolor{gray!10}{59.39 (7.5)} & \cellcolor{gray!10}{11.35 (0.12)}\\
\hspace{1em}RFSRC & 71.6 (2.3) & 9.81 (0.33)\\
\hspace{1em}\cellcolor{gray!10}{RFSRC\_DT} & \cellcolor{gray!10}{71.29 (2.13)} & \cellcolor{gray!10}{9.84 (0.28)}\\
\hspace{1em}XGBCox & 71.18 (2.24) & 9.87 (0.29)\\
\hspace{1em}\cellcolor{gray!10}{XGB\_PEM} & \cellcolor{gray!10}{71.2 (2.62)} & \cellcolor{gray!10}{9.83 (0.35)}\\
\hspace{1em}XGB\_DT & 70.79 (2.58) & 9.91 (0.3)\\
\addlinespace[0.3em]
\multicolumn{3}{l}{\textbf{tumor}}\\
\hspace{1em}\cellcolor{gray!10}{KM} & \cellcolor{gray!10}{50 (0)} & \cellcolor{gray!10}{20.06 (0.56)}\\
\hspace{1em}RIDGE & 63.9 (1.84) & 20.06 (0.56)\\
\hspace{1em}\cellcolor{gray!10}{GLMN} & \cellcolor{gray!10}{50 (0)} & \cellcolor{gray!10}{20.06 (0.56)}\\
\hspace{1em}RFSRC & 63.61 (1.57) & 19.07 (0.5)\\
\hspace{1em}\cellcolor{gray!10}{RFSRC\_DT} & \cellcolor{gray!10}{61.42 (1.89)} & \cellcolor{gray!10}{19.63 (0.73)}\\
\hspace{1em}XGBCox & 60.05 (3.58) & 20.43 (1.44)\\
\hspace{1em}\cellcolor{gray!10}{XGB\_PEM} & \cellcolor{gray!10}{62.33 (1.52)} & \cellcolor{gray!10}{19.16 (0.93)}\\
\hspace{1em}XGB\_DT & 61.17 (2.79) & 19.2 (0.85)\\*
\end{longtable}

\begin{figure}[ht]
    \centering
    \includegraphics[width=0.9\linewidth]{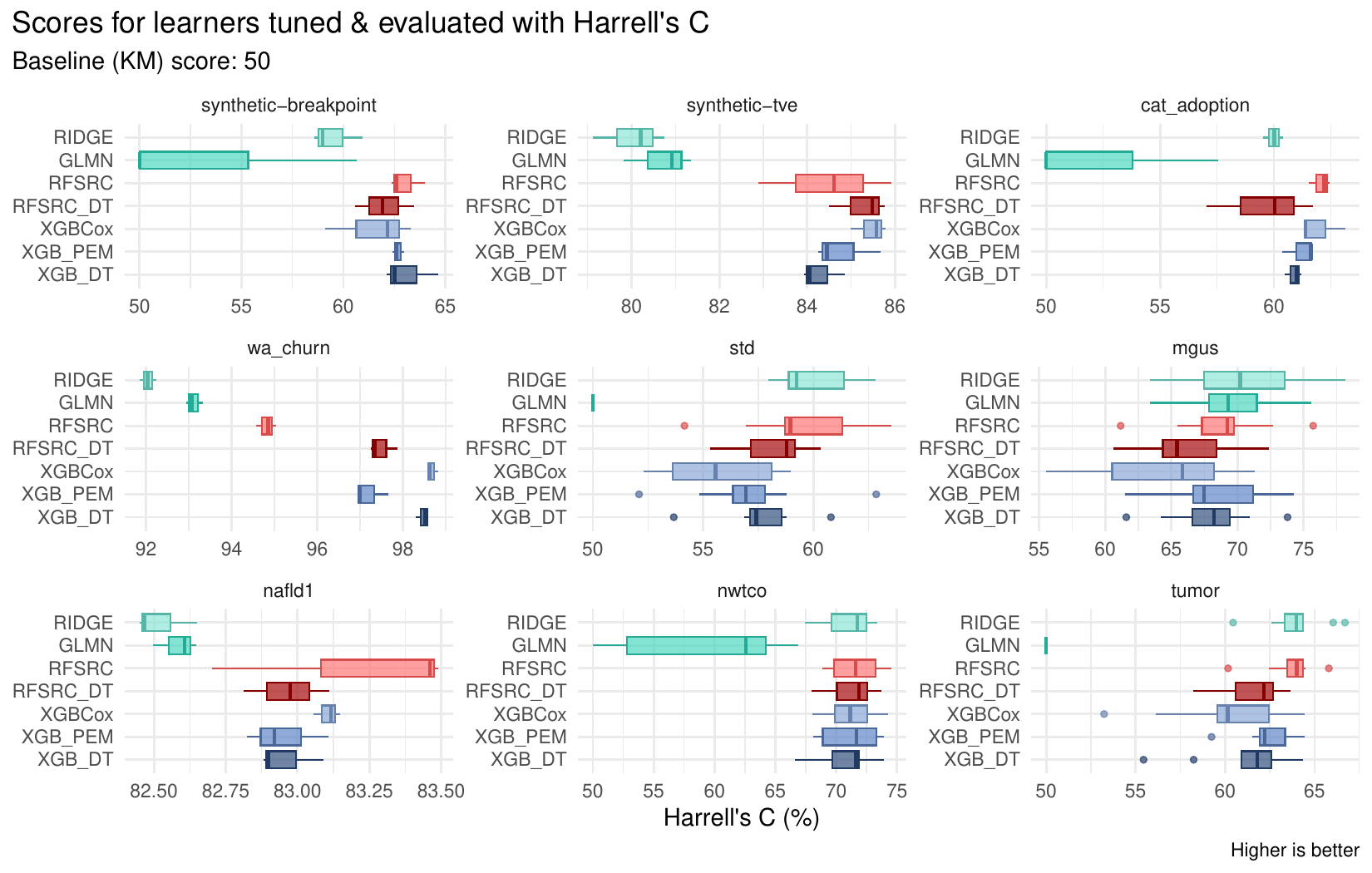}
    \caption{Boxplots of Harrell's C evaluation scores across each task's outer resampling iterations, scaled by 100. Baseline (KM) scores are omitted as they are equal to 50 in all cases.}
    \label{fig:bm-scores-harrell-c}
\end{figure}

\begin{figure}[ht]
    \centering
    \includegraphics[width=0.9\linewidth]{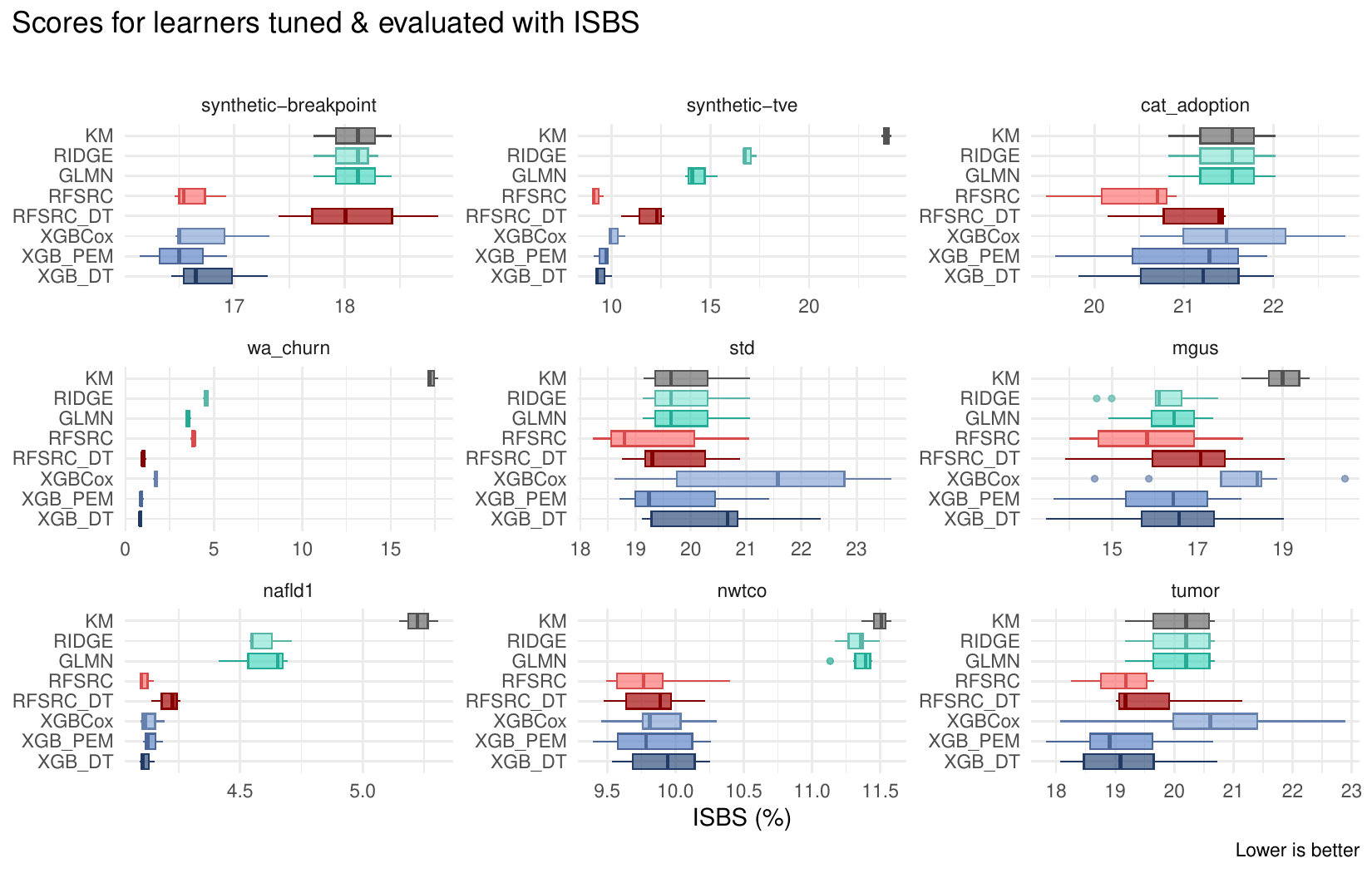}
    \caption{Boxplots of integrated survival Brier scores (ISBS) scores across each task's outer resampling iterations, scaled by 100.}
    \label{fig:bm-scores-isbs}
\end{figure}

\clearpage
\bibliographystyle{plainnat}
\bibliography{main, pem, discrete, pseudo}

\end{document}